\documentclass{article}

\usepackage{arxiv}

\usepackage[utf8]{inputenc} % allow utf-8 input
\usepackage[T1]{fontenc}    % use 8-bit T1 fonts
\usepackage{hyperref}       % hyperlinks
\usepackage{url}            % simple URL typesetting
\usepackage{booktabs}       % professional-quality tables
\usepackage{amsfonts}       % blackboard math symbols
\usepackage{nicefrac}       % compact symbols for 1/2, etc.
\usepackage{microtype}      % microtypography
\usepackage{lipsum}
\usepackage{appendix}
\usepackage{graphicx}
\usepackage{adjustbox}
\usepackage{array}
\usepackage{float}
\usepackage{subcaption} 
\graphicspath{ {./images/} }

\title{Few-Shot Remote Sensing Image Scene Classification with CLIP and Prompt Learning}

\author{
 Ivica Dimitrovski \\
  Faculty of Computer Science and Engineering\\
  University Ss Cyril and Methodius\\
  Skopje 1000, North Macedonia \\
  \texttt{ivica.dimitrovski@finki.ukim.mk} \\
   \And
 Vlatko Spasev \\
  Faculty of Computer Science and Engineering\\
  University Ss Cyril and Methodius\\
  Skopje 1000, North Macedonia \\
  \texttt{vlatko.spasev@finki.ukim.mk} \\
  \And
 Ivan Kitanovski \\
  Faculty of Computer Science and Engineering\\
  University Ss Cyril and Methodius\\
  Skopje 1000, North Macedonia \\
  \texttt{ivan.kitanovski@finki.ukim.mk} \\
}

\begin{document}
\maketitle
\begin{abstract}
Remote sensing applications increasingly rely on deep learning for scene classification. However, their performance is often constrained by the scarcity of labeled data and the high cost of annotation across diverse geographic and sensor domains. While recent vision-language models (VLMs) like Contrastive Language-Image Pre-Training (CLIP) have shown promise by learning transferable representations at scale by aligning visual and textual modalities, their direct application to remote sensing remains suboptimal due to significant domain gaps and the need for task-specific semantic adaptation. To address this critical challenge, we systematically explore prompt learning as a lightweight and efficient adaptation strategy for few-shot remote sensing image scene classification. We evaluate several representative methods, including Context Optimization (CoOp), Conditional Context Optimization (CoCoOp), Multi-modal Prompt Learning (MaPLe), and Prompting with Self-Regulating Constraints (PromptSRC). These approaches reflect complementary design philosophies: from static context optimization to conditional prompts for enhanced generalization, multi-modal prompts for joint vision-language adaptation, and semantically regularized prompts for stable learning without forgetting. We benchmark these prompt-learning methods against two standard baselines: zero-shot CLIP with hand-crafted prompts and a linear probe trained on frozen CLIP features. Through extensive experiments on multiple benchmark remote sensing datasets, including cross-dataset generalization tests, we demonstrate that prompt learning consistently outperforms both baselines in few-shot scenarios. Notably, PromptSRC achieves the most robust cross-domain performance. Our findings underscore prompt learning as a scalable and efficient solution for bridging the domain gap in satellite and aerial imagery, providing a strong foundation for future research in this field. 
\end{abstract}

% keywords can be removed

\section{Introduction}
Remote sensing image scene classification is a fundamental task in Earth observation, enabling automated understanding of land use and land cover from aerial or satellite images \cite{DIMITROVSKI202318}, \cite{Cheng2017_RSSurvey}. Traditional approaches rely on supervised learning techniques that require large volumes of manually labeled data \cite{Cheng8252784}, \cite{Cheng9127795}. However, acquiring annotated remote sensing data is both time-consuming and costly, especially across diverse geographic regions, seasons, and imaging modalities. Moreover, such learning approaches limit visual recognition systems to a fixed set of known classes. To handle new classes, the system requires additional data and retraining. These limitations have motivated a growing interest in the development of label-efficient or few-shot learning methods for remote sensing applications \cite{qiu2024few}, \cite{sun2021research}, \cite{10401319}. 

Few-shot learning aims to enable models to recognize new classes using only a limited number of labeled examples, a setting that reflects the data scarcity often encountered in remote sensing. Several learning paradigms have been proposed to address this challenge. Meta-learning, or “learning to learn,” trains models across a variety of tasks so they can rapidly adapt to new ones with minimal supervision, capturing transferable strategies rather than task-specific features. Transfer learning follows a complementary approach, leveraging pre-training on large, diverse datasets and fine-tuning on smaller, domain-specific ones to reduce annotation costs while maintaining strong performance. In addition, metric-based approaches such as Siamese Networks \cite{koch2015siamese}, Prototypical Networks \cite{snell2017prototypical}, and Matching Networks \cite{vinyals2016matching} focus on learning robust similarity measures between samples, enabling models to classify unseen categories by comparing them to a small support set of known examples. Together, these methods have significantly advanced few-shot recognition but still rely on task-specific training and limited generalization across domains. 

This has motivated recent interest in vision-language models that learn transferable, semantically grounded representations capable of supporting few-shot and even zero-shot learning without explicit retraining \cite{LI2023103497}. Among these, recent vision-language models (VLMs), such as CLIP \cite{radford2021learning}, BLIP \cite{li2022blip}, and ALIGN \cite{jia2021scaling} have demonstrated remarkable capabilities in visual representation learning. These models are trained on large-scale datasets comprising hundreds of millions to billions of image-text pairs, enabling them to learn a shared embedding space where visual and textual inputs are semantically aligned. A key strength of these models lies in their ability to perform zero-shot classification: instead of requiring labeled training data for each target class, they can directly associate an input image with natural language descriptions of class categories. This facilitates open-vocabulary recognition without any task-specific fine-tuning, making them particularly attractive for scenarios where labeled data is scarce or constantly evolving, such as remote sensing.

Despite the strong generalization capabilities of vision-language models, their direct application to remote sensing remains non-trivial \cite{10506064}, \cite{jain2025senclip}. The domain shift between natural images, on which these models are pre-trained, and remote sensing imagery introduces challenges related to differences in viewpoint, scale, texture, and semantic content. Moreover, textual descriptions commonly used in VLMs training may not align well with the specialized terminology and class structure typical in remote sensing tasks. As a result, zero-shot performance often degrades when applied naively to aerial or satellite imagery. These limitations underscore the need for domain adaptation techniques that can bridge the semantic and distributional gap between the source (natural image) and target (remote sensing) domains, while preserving the efficiency and flexibility of pre-trained models. Among the available strategies, prompt learning has emerged as a promising solution for aligning vision-language models to new tasks without retraining the entire network \cite{shin2020autoprompt}.

In this work, we explore the adaptation of the CLIP vision-language model to remote sensing through the lens of prompt learning. Specifically, we investigate a series of prompt-learning strategies that enable efficient and scalable fine-tuning of CLIP without modifying its pre-trained backbone or requiring large-scale domain-specific pretraining. Starting from the foundational Context Optimization (CoOp) approach \cite{zhou2022learning}, which learns continuous prompt embeddings from limited labeled samples, we extend our analysis to three advanced variants: Conditional Context Optimization (CoCoOp) \cite{zhou2022conditional}, Multi-modal Prompt Learning (MaPLe) \cite{khattak2023maple}, and Prompting with Self-Regulating Constraints (PromptSRC) \cite{khattak2023self}. These methods progressively enhance the adaptability of CLIP by incorporating instance-conditioned prompts, cross-modal coupling between vision and language representations, and self-regularization mechanisms that improve stability and generalization.

We benchmark these prompt-learning techniques against two representative baselines: (i) zero-shot CLIP using carefully handcrafted textual prompts, and (ii) a linear probe that trains a shallow classifier on top of CLIP’s frozen image encoder. Our experiments span multiple benchmark remote sensing datasets, evaluating performance under few-shot and cross-dataset transfer settings to assess data efficiency, robustness, and domain generalization.

Our results show that prompt learning substantially improves CLIP’s transferability to remote sensing imagery. In particular, CoOp and CoCoOp provide strong few-shot adaptation, MaPLe achieves improved cross-modal alignment, and PromptSRC further enhances robustness through self-regulating constraints. Collectively, these findings highlight the potential of prompt-learning paradigms as lightweight, scalable, and generalizable adaptation strategies for foundation models in Earth observation. This work contributes to the growing body of research advocating for the use of vision-language models in remote sensing and provides new insights into how prompt-based methods can bridge the semantic and distributional gaps between natural and domain-specific imagery.

This work makes the following contributions to few-shot remote sensing image scene classification with vision-language models:
\begin{itemize}
    \item \textit{Unified benchmark of prompt learning for remote sensing.} We present a controlled, reproducible evaluation of four representative prompt-learning paradigms: CoOp, CoCoOp, MaPLe, and PromptSRC, built on CLIP (ViT-B/16), under strict $k$-shot regimes ($k\!\in\!\{1,2,4,8,16\}$) across nine widely used datasets, using standardized splits and identical pre-processing.
    \item \textit{Lightweight, architecture-agnostic adaptation pipeline.} We freeze the CLIP backbone and adapt only prompt parameters, providing a practical recipe (hyper-parameters, templates, training schedule) that achieves strong performance on a single commodity GPU without domain-specific pre-training.
    \item \textit{Consistent gains over strong baselines.} Across all datasets and shot counts, prompt learning consistently outperforms both zero-shot CLIP (hand-crafted prompts) and a linear probe on frozen CLIP features, validating prompts as an efficient alternative to classifier retraining in data-scarce settings.
    \item \textit{Cross-dataset generalization at scale.} We perform a comprehensive source-target transfer study covering all dataset pairs, revealing robust cross-domain behavior for PromptSRC and strong performance of MaPLe on heterogeneous pairs.
    \item \textit{Diagnostic analysis for actionable insights.} Beyond overall accuracy, we provide per-dataset normalized confusion matrices and qualitative error analysis that pinpoint systematic confusions (e.g., semantically adjacent urban and vegetation classes), suggesting avenues such as hierarchical or context-aware prompting.
\end{itemize}

The remainder of this paper is structured as follows. Section~2 reviews related work and provides background on the application of vision-language models in remote sensing. Section~3 presents the datasets and prompt-learning methods evaluated in this study, including CoOp, CoCoOp, MaPLe, and PromptSRC. Section~4 outlines the experimental design, evaluation protocols, and implementation details. Section~5 reports and analyzes the results across multiple remote sensing benchmarks. Finally, Section~6 concludes the paper and discusses future research directions.

\section{Related work and background}
Scene classification in remote sensing has been extensively explored, particularly under the supervised learning paradigm, where large annotated datasets enable the training of deep models with high accuracy~\cite{DIMITROVSKI202318}. Early work in remote sensing image scene classification relied on hand-crafted features such as texture, spectral, and spatial descriptors, combined with classical machine learning classifiers~\cite{cheng2017remote}. With the advent of deep learning, convolutional neural networks (CNNs) became the dominant paradigm, significantly improving performance when abundant labeled data is available. However, in practical applications, obtaining large quantities of labeled data remains difficult, especially for novel or fine-grained scene classes. This limitation has motivated the shift toward few-shot learning approaches, which aim to achieve robust generalization from only a few labeled samples per class.

A major branch of few-shot learning is metric-based learning, which learns an embedding space where similarity between samples can be measured through metrics like Euclidean or cosine distance. In remote sensing, where inter-class similarity is high and intra-class variability is large, simple Euclidean distances are often inadequate. To overcome this, several specialized approaches have been developed. For instance, the Multi-scale Covariance Metric Network (MCMNet) maps deep features to a Riemannian manifold space using covariance representations, yielding more robust similarity estimates for high-resolution data~\cite{chen2023few}. Similarly, RS-MetaNet introduces meta-metric learning, organizing training into a family of few-shot tasks to learn a transferable metric space for remote sensing scenes~\cite{li2020rs}. The recently proposed Masked Second-Order Pooling (MSoP) framework~\cite{deng2023masked} enhances feature representations by modeling second-order statistics and selectively masking dominant regions, effectively mitigating intra-class and inter-class similarity issues in few-shot remote sensing scene classification. In a related direction, Parameter-free Attention and Region Matching (PARM)~\cite{jia2025few} introduces an attention-guided region matching strategy that captures local spatial correspondences without learnable parameters, further improving discriminative power under data-scarce conditions.

Beyond metric learning, meta-learning or "learning to learn" strategies train a meta-learner capable of rapidly adapting model parameters to new tasks with minimal updates. The Scene Graph Matching Network (SGMNet), for example, explicitly models object co-occurrence and spatial relationships within remote sensing scenes, using graph structures to perform few-shot classification~\cite{zhang2022sgmnet}. Meanwhile, HCPNet focuses on learning discriminative prototypes by generating query-specific class boundaries, improving separation between visually similar classes under few-shot conditions~\cite{zhu2023hcpnet}.

Complementary to these methods, self-supervised and augmentation-based techniques leverage large unlabeled datasets or enhanced data diversity to improve few-shot generalization. Alosaimi \textit{et al.}~\cite{alosaimi2023self} employ self-supervised pretraining on unlabeled remote sensing imagery using a dual-branch network that processes both low- and high-resolution images, significantly improving performance in low-shot regimes. Similarly, Dong \textit{et al.}~\cite{dong2024optimizing} propose a data augmentation framework that diversifies the support set through controlled distortions and a dual-path classification strategy, enhancing robustness in few-shot scenarios.

Another research direction explores feature fusion and hybrid methods, combining multiple representations to enhance discriminative power under limited data conditions. Yang \textit{et al.}~\cite{yang2023few} integrate CNN-derived features with discrete wavelet transform (DWT) subband features to form multi-subband representations, followed by discriminant correlation analysis to mitigate inter-class confusion in 1-2 shot settings. More recent efforts also integrate ensemble-based self-supervised pretraining and task-specific contrastive learning to further improve generalization in remote sensing few-shot learning \cite{li2023multiform}, \cite{ma2023multipretext}.

In parallel, recent advances in vision-language models (VLMs) have opened new opportunities for few-shot and zero-shot remote sensing classification. While models such as CLIP demonstrate remarkable generalization in natural image domains, transferring them to aerial and satellite imagery presents challenges due to the unique characteristics of remote sensing data top-down geometry, high inter-class similarity, and substantial domain shifts from natural image distributions. To bridge this gap, several domain-adapted VLMs have been proposed. Models such as RemoteCLIP~\cite{liu2024remoteclip}, RS-CLIP~\cite{li2023rs}, SkyCLIP~\cite{wang2024skyscript}, GeoRSCLIP~\cite{zhang2306rs5m}, and SenCLIP~\cite{jain2025senclip} extend the CLIP framework with remote sensing specific datasets and fine-tuning strategies to better align visual and textual representations. Recent efforts have also extended CLIP adaptation beyond classification tasks. For instance, Liu \textit{et al.}~\cite{liu2024clip} propose a CLIP-guided source-free object detection framework for aerial images, demonstrating the versatility of vision-language alignment for cross-domain transfer, even outside scene classification.

Each of these models introduces complementary innovations. RemoteCLIP continually pre-trains on a collection of remote sensing datasets covering detection, segmentation, and image-text tasks, effectively specializing CLIP for aerial imagery. RS-CLIP employs pseudo-labeling and curriculum learning to progressively adapt to domain-specific data without extensive manual annotation. SkyCLIP leverages OpenStreetMap data to automatically pair satellite images with semantic descriptions, demonstrating strong performance in zero-shot and fine-grained tasks. GeoRSCLIP integrates geospatial grounding by linking millions of images with geotagged textual descriptions, while SenCLIP enhances cross-view understanding by associating Sentinel-2 satellite imagery with ground-level photographs, reducing domain mismatches between modalities. Related cross-view models, such as Sat2Cap~\cite{dhakal2024sat2cap}, further improve semantic grounding by linking overhead imagery with descriptive ground-level captions.

In addition to model-level adaptation, prompt engineering has proven to be an effective means of improving CLIP-based classification for remote sensing. Zhang \textit{et al.}~\cite{app132212462} show that structured, domain-specific prompts incorporating cues such as spatial scale or land-cover type significantly enhance zero-shot accuracy across multiple benchmarks. They also demonstrate that prompt ensembling and fine-grained textual templates can reduce ambiguity and better align language with visual content, underscoring the importance of prompt formulation in remote sensing VLMs.

Building on this insight, prompt learning has emerged as a powerful, parameter-efficient adaptation strategy~\cite{cui2025enhancing}. Rather than modifying model weights or requiring large-scale pretraining, prompt learning optimizes the input prompts themselves. Foundational methods such as Context Optimization (CoOp)~\cite{zhou2022learning} learn continuous prompt embeddings from limited samples, while Conditional CoOp (CoCoOp)~\cite{zhou2022conditional} introduces instance-dependent prompts to improve generalization. Later frameworks like MaPLe~\cite{khattak2023maple} jointly adapt vision and language branches through cross-modal coupling, and PromptSRC~\cite{khattak2023self} incorporate self-regularization to balance adaptability and generalization. Collectively, these approaches demonstrate the growing maturity of prompt learning as a lightweight yet effective paradigm for tailoring foundation models to the unique challenges of remote sensing, particularly under few-shot and data-scarce conditions.

Despite notable advances in few-shot learning and the emergence of domain-adapted vision-language models for remote sensing, current approaches often depend on extensive re-training or large-scale domain-specific datasets to achieve satisfactory performance. Such requirements limit their scalability, efficiency, and adaptability to dynamic or data-scarce environments. In this context, our work positions prompt learning as a lightweight and architecture-agnostic alternative for adapting foundation vision-language models to remote sensing imagery. By systematically evaluating representative paradigms including CoOp, CoCoOp, MaPLe, and PromptSRC across diverse datasets and few-shot regimes, we provide a unified and reproducible benchmark that elucidates their relative strengths in terms of adaptability, generalization, and robustness. Through this comprehensive analysis, we highlight prompt learning as a practical and effective strategy to achieving label-efficient, transferable, and semantically grounded scene understanding in Earth observation.

%%%%%%%%%%%%%%%%%%%%%%%%%%%%%%%%%%%%%%%%%%
\section{Data and methods}
This section presents the datasets and methodological framework used in our study. We first describe the nine benchmark remote sensing datasets employed to evaluate the adaptability of vision-language models across varying spatial resolutions, scene types, and acquisition modalities. Subsequently, we review the CLIP model architecture and its zero-shot inference mechanism, which form the foundation of our experimental setup. Building upon this foundation, we introduce a series of prompt learning strategies: CoOp, CoCoOp, MaPLe, and PromptSRC. Each method is discussed in detail, accompanied by a unified comparative visualization presented in Figure~\ref{fig:prompt-learning}, and analyzed in the context of its relevance and applicability to Earth observation imagery.

\subsection{Data description}
With the growing volume of remote sensing imagery, the research community has invested significant effort in developing and sharing high-quality annotated datasets to advance machine learning in Earth observation \cite{DIMITROVSKI202318}, \cite{spasev2023semantic}. In this study, we compile a benchmark consisting of nine widely used and publicly available remote sensing datasets. These datasets vary substantially in terms of size, spatial resolution, image format, and sensor modality. This variety provides a robust foundation for evaluating the adaptability of vision-language models in remote sensing scenarios.

\begin{table*}[!ht]
\centering
\caption{Summary of the remote sensing image scene classification datasets.}
\begin{adjustbox}{width=0.89\linewidth}
\begin{tabular}{lllllllll}
\hline
\textbf{Name} & \textbf{\rotatebox[origin=c]{90}{Image type}} & \textbf{\rotatebox[origin=c]{90}{\#Images}} & \textbf{\rotatebox[origin=c]{90}{Image size}} & \textbf{\rotatebox[origin=c]{90}{Spatial resolution}} & \textbf{\rotatebox[origin=c]{90}{\#Labels}} \\
\hline
UC Merced \cite{yang2010uc_merced} & Aerial RGB & 2100    & 256×256 & 0.3m    & 21  \\
\hline
AID \cite{xia2017aid}  & Aerial RGB & 10000   & 600×600 & 0.5m - 8m    & 30 \\
\hline
EuroSAT \cite{helber2019eurosat}   & Sat. Multispectral      & 27000   & 64×64 & 10m      & 10  \\
\hline
RESISC45 \cite{Cheng2017_RSSurvey}  & Aerial RGB & 31500   & 256×256 & 0.2m - 30m    & 45  \\
\hline
RSSCN7 \cite{Zou2015RSSCN7}    & Aerial RGB & 2800    & 400×400 & n/a   & 7 \\
\hline
SIRI-WHU \cite{zhu2016siriwhu}  & Aerial RGB & 2400    & 200×200 & 2m    & 12 \\
\hline
CLRS \cite{haifeng2020clrs} & Aerial RGB & 15000   & 256×256  & 0.26m - 8.85m   & 25 \\
\hline
Optimal-31 \cite{qi2019optimal31} & Aerial RGB & 1860  & 256×256   & n/a & 31  \\
\hline
MLRSNet \cite{qi2020mlrsnet}     & Aerial RGB  & 109161       & 256×256 & 0.1m - 10m        & 47    \\
\hline
\end{tabular}
\end{adjustbox}
\label{table:datasets}
\end{table*}

Table~\ref{table:datasets} summarizes the characteristics of the selected remote sensing datasets. The number of images varies substantially across datasets, ranging from approximately 1,000 to nearly 110,000. Similarly, the number of class labels ranges from 7 to 47, reflecting the diversity of classification complexity across tasks. Most datasets consist of aerial RGB images, while EuroSAT is originally a multispectral dataset based on 13 Sentinel-2 bands. However, for our experiments, we use the official RGB version provided by the authors \cite{helber2019eurosat}, constructed from the B4 (red), B3 (green), and B2 (blue) bands. The datasets also vary significantly in terms of spatial resolution and image size.

\begin{figure}[H]
\centering
  \includegraphics[trim={0 0 0 0},clip, width=1.0\linewidth]{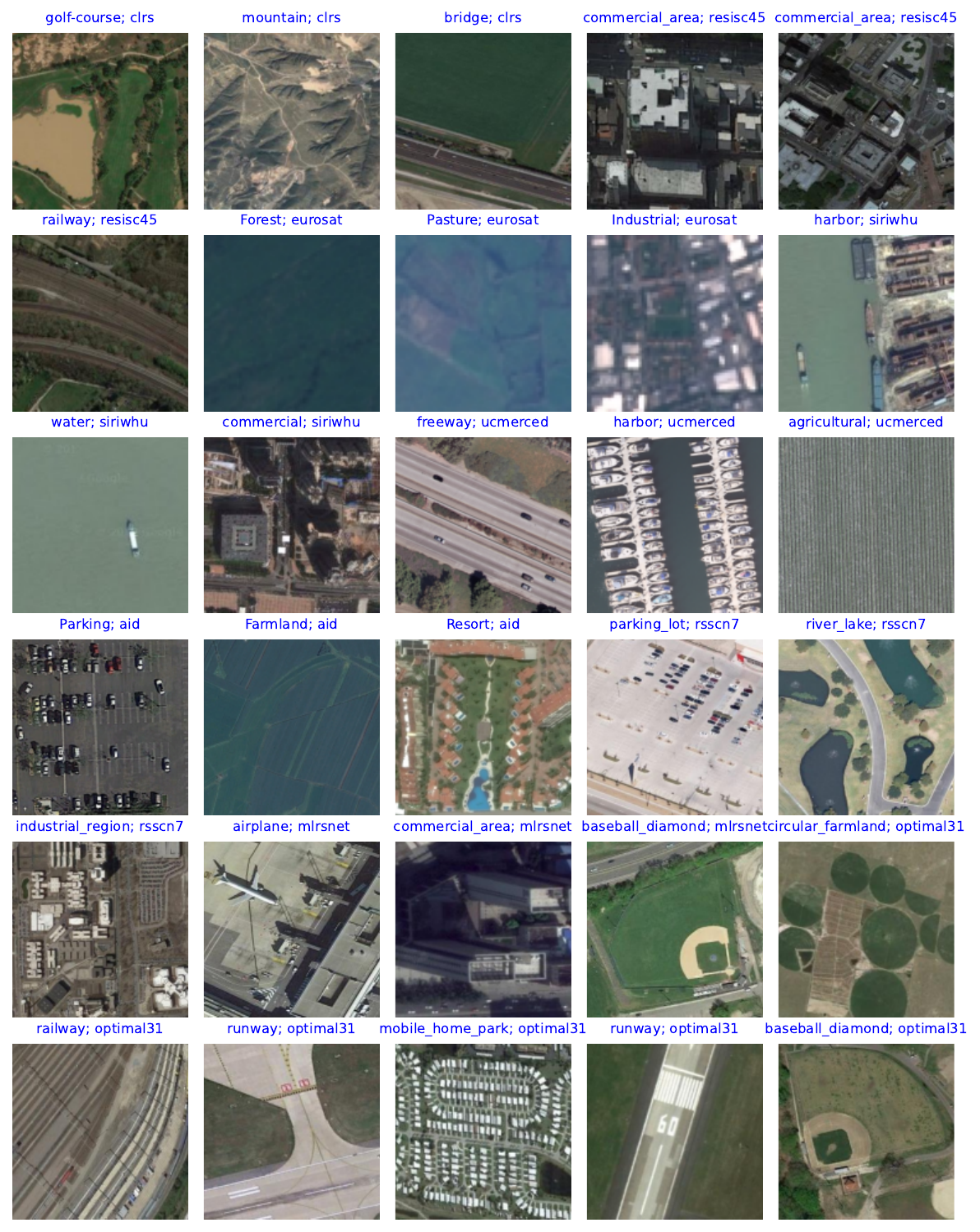}
  \caption{Illustrative examples from the datasets used in our study, showing the dataset source and the associated class label for each image.}
  \label{fig:example_images}
\end{figure}

To illustrate the diversity across datasets, Figure~\ref{fig:example_images} presents representative sample images of each of the nine datasets. These examples highlight the significant variation in spatial resolution, scene structure, and color distribution. By including both high-resolution aerial imagery and coarser satellite views, as well as rural, urban, and industrial landscapes, the figure underscores the challenges of generalizing across heterogeneous remote sensing data. This visual overview supports our motivation for using a broad benchmark to evaluate the robustness of CLIP-based models and adaptation strategies such as prompt learning.

\subsection{Review of CLIP}
Contrastive Language-Image Pre-Training (CLIP), introduced by OpenAI \cite{radford2021learning}, is a vision-language foundation model that learns to connect images and natural language descriptions within a shared embedding space. CLIP consists of two separate encoders: an image encoder and a text encoder. The image encoder is typically a Vision Transformer (ViT) \cite{dosovitskiy2020image} or ResNet \cite{he2016deep}. The text encoder is based on a Transformer architecture \cite{vaswani2017attention}, incorporating modifications described in \cite{radford2019language}, and configured as a base model with 63 million parameters, consisting of 12 layers, a hidden width of 512, and 8 attention heads. Each encoder maps its respective modality into a joint multi-modal embedding space, where similarity can be computed using the cosine distance between normalized embeddings. This dual-encoder design allows CLIP to process images and text independently, making it computationally efficient at inference time and scalable to large datasets.

CLIP is trained using a contrastive learning objective on a dataset containing 400 million image-text pairs collected from the Internet. The goal is to bring the embeddings of matched image-text pairs closer together while separating the mismatched pairs. The training process involves computing a similarity matrix between all image and text embeddings in a batch and applying a symmetric cross-entropy loss over the image-to-text and text-to-image pairs. This encourages the model to associate each image with its correct textual description and vice versa, effectively aligning the visual and textual modalities.

\subsection{Prompting and zero-shot inference}
CLIP possesses several distinctive characteristics that make it particularly effective for open-ended visual recognition tasks. First, its training is based on contrastive learning over image-text pairs rather than conventional classification with predefined category labels. This approach allows CLIP to leverage massive amounts of freely available data from the web, such as social media captions and alt-text, without the need for costly manual annotation. Second, the scale and diversity of the training corpus equip CLIP with a robust understanding of general visual and textual concepts. Third, since the textual descriptions often reference multiple attributes or contextual elements within an image, the model learns to build holistic representations that go beyond narrow, task-specific features.

\begin{figure}
\centering
  \includegraphics[trim={0 0 0 0},clip, width=1.0\linewidth]{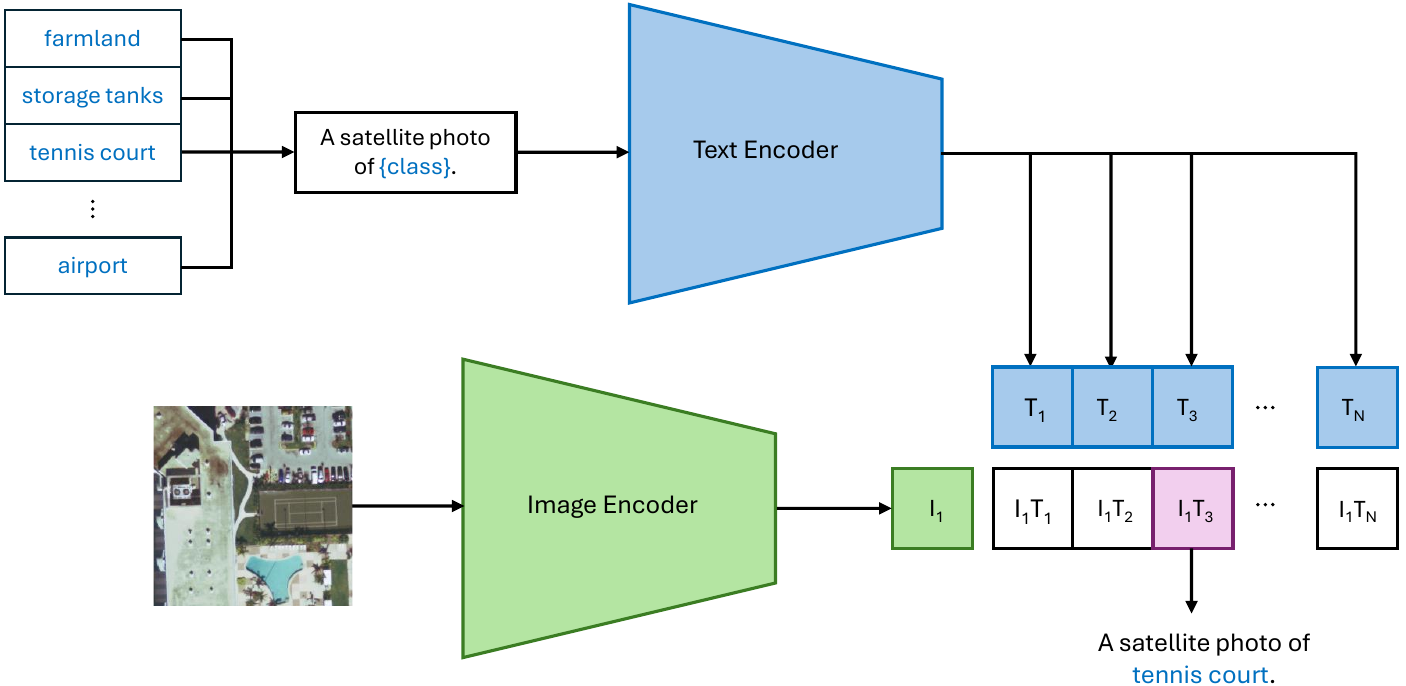}
  \caption{Illustration of zero-shot classification using CLIP. Class labels are reformulated as natural language prompts and encoded into text embeddings. A satellite image of a tennis court is encoded into an image embedding and compared to all text embeddings using cosine similarity. The correct prediction corresponds to the prompt with the highest similarity, e.g., “a satellite photo of a tennis court”.}
  \label{fig:zero_shot_inference}
\end{figure}

These characteristics collectively enable CLIP to perform zero-shot inference through prompt-based classification. Zero-shot image classification refers to the task of assigning semantic classes to images without having seen labeled examples from those specific classes during training \cite{wang2019survey}. As illustrated in Figure~\ref{fig:zero_shot_inference}, the process begins by encoding a set of textual prompts representing candidate classes (e.g., “a satellite image of a forest”, “an aerial image of an airport”) into text embeddings. Simultaneously, the input image is passed through the image encoder to generate a corresponding image embedding. The model then computes the cosine similarity between the image embedding and each text embedding to measure their semantic alignment. Cosine similarity quantifies the angle between two vectors in the embedding space, with higher values indicating greater similarity. The class associated with the text prompt that yields the highest similarity score is selected as the predicted label. This inference process enables CLIP to recognize categories it has never explicitly encountered during training, relying solely on the semantic content embedded in the natural language descriptions. This approach is particularly useful in open-set recognition problems, where the set of target categories may change or grow over time.

A key factor influencing performance is prompt engineering: the design of textual descriptions for each class. Well-crafted prompts that match the linguistic patterns seen in CLIP’s pre-training data can significantly improve accuracy. In remote sensing, this might involve specifying both the modality and the context, e.g., “a satellite photo of tennis court” rather than simply “tennis court.” However, manually designing optimal prompts can be labor-intensive and may still fail to capture domain-specific semantics.

CLIP’s zero-shot capabilities have demonstrated remarkable generalization across a wide variety of domains \cite{radford2021learning}, including remote sensing \cite{app132212462}, where domain shifts from natural to overhead imagery can challenge traditional classifiers. When paired with effective prompt engineering strategies and high-capacity vision backbones, CLIP serves as a powerful foundation model for flexible, label-efficient classification pipelines.

\subsection{Context optimization}
Context Optimization (CoOp) introduced the foundational idea of learning continuous textual prompts as a lightweight mechanism for adapting vision-language models to downstream tasks~\cite{zhou2022learning}. Instead of relying on handcrafted textual templates such as "a satellite photo of an airport", CoOp optimizes a set of learnable context tokens that replace or augment the manual prompt, allowing the model to automatically discover discriminative task-specific textual cues. An overview of the CoOp architecture and its relation to the frozen CLIP encoders is illustrated in Figure~\ref{fig:coop}.

In CLIP, each text prompt is first tokenized using byte pair encoding (BPE)~\cite{sennrich2015neural}, which converts every token (including punctuation) into a unique numeric identifier drawn from a 49,152-word vocabulary. For batch processing, the token sequence is prepended with a start-of-sequence ([SOS]) token and appended with an end-of-sequence ([EOS]) token, then padded or truncated to a fixed length of 77 tokens. These token IDs are embedded into 512-dimensional word vectors and passed through CLIP’s Transformer-based text encoder. The hidden state corresponding to the [EOS] position is then layer-normalized and linearly projected to obtain the final text embedding within CLIP’s shared vision-language feature space.

CoOp modifies this pipeline by replacing a portion of the fixed textual tokens with learnable context embeddings that are optimized during training. These continuous vectors, when concatenated with the tokenized class name (e.g., "airport", "farmland"), form the final prompt fed into the frozen text encoder. The model thus learns how to represent contextual modifiers that maximize alignment between text and image embeddings. Importantly, only the prompt embeddings are updated, while the CLIP encoders remain frozen, preserving the generalization capacity of the pre-trained model.

This approach enables parameter-efficient adaptation while maintaining the integrity of the vision-language embedding space. In remote sensing applications, CoOp allows CLIP to internalize domain-specific textual contexts that better describe aerial imagery, such as "a remote sensing image of a highway" or "a satellite photo of farmland". Consequently, CoOp effectively bridges handcrafted and learned prompts, achieving robust performance even when annotated samples are scarce.

\subsection{Conditional context optimization}
Conditional Context Optimization (CoCoOp) builds directly upon the CoOp framework and retains the same textual encoding pipeline, where prompts are tokenized using byte pair encoding (BPE) and passed through CLIP’s frozen Transformer-based text encoder~\cite{zhou2022conditional}. However, unlike CoOp which learns a fixed, shared set of prompt embeddings for all samples, CoCoOp introduces a conditional mechanism that adapts the prompt representation dynamically for each input image. This modification addresses CoOp’s limitation in generalizing to unseen classes or domains by allowing the textual context to vary based on the visual characteristics of the query image. The overall architecture and conditional prompt generation mechanism are illustrated in Figure~\ref{fig:cocoop}.

The key innovation in CoCoOp is the integration of a lightweight meta-network that generates instance-dependent prompt vectors conditioned on the image features extracted by CLIP’s frozen visual encoder. These dynamically generated prompts replace the static textual context used in CoOp, enabling the model to flexibly adapt its semantic representation to the specific appearance or structure of each visual input. This mechanism effectively establishes a feedback connection between the vision and language branches while maintaining the efficiency and stability of CLIP’s frozen backbone.

Through this conditional adaptation, CoCoOp significantly improves base-to-novel class generalization, particularly in scenarios where high intra-class variation or visual ambiguity exists. For remote sensing image scene classification, this design allows the model to adapt to diverse imaging conditions such as seasonal changes, illumination differences, or varying spatial resolutions. By tailoring prompts to each visual instance, CoCoOp enhances recognition robustness in complex Earth observation datasets where static textual contexts may fail to capture the full variability of scene appearance.

\subsection{Multi-modal prompt learning}
Multi-modal prompt learning extends CLIP-based prompt tuning by jointly optimizing prompts in both the visual and textual branches. This dual adaptation strengthens cross-modal alignment and enhances generalization across diverse domains~\cite{khattak2023maple}. Unlike CoOp and CoCoOp, which adapt only the textual side of CLIP, MaPLe introduces a unified framework that integrates deep hierarchical prompting with explicit vision-language coupling. A schematic of MaPLe’s dual-branch, layer-wise prompting and coupling strategy is shown in Fig.~\ref{fig:maple}.

In MaPLe, each transformer layer within both the image and text encoders is augmented with a small set of learnable prompt tokens that serve as contextual embeddings. These prompts are inserted alongside the standard token sequences, patch tokens in the vision branch and word tokens in the text branch, allowing the model to refine representations progressively at multiple abstraction levels. This deep prompting strategy enables more flexible and expressive adaptation than single-layer or shallow prompt tuning.

A central innovation of MaPLe is its coupling mechanism, which projects the text prompts into the visual-prompt space to maintain semantic coherence between modalities. This coupling ensures that prompts in both branches evolve jointly rather than independently, thereby reinforcing vision-language alignment during adaptation. The coupling is lightweight and learnable, allowing the model to share contextual information across encoders without altering the frozen CLIP backbone.

During training, only the prompt tokens and coupling parameters are optimized, while all pre-trained CLIP weights remain fixed. This design preserves the strong generalization properties of the foundation model while enabling efficient domain adaptation with minimal additional parameters. In remote sensing applications, multi-modal prompting provides improved robustness to domain shifts arising from differences in spatial resolution, atmospheric conditions, and imaging geometry, scenarios where uni-modal prompt learning often underperforms.

\subsection{Prompting with self-regulating constraints}
\begin{figure*}[t]
    \centering

    % ---------- Row 1 ----------
    \begin{subfigure}[t]{0.45\textwidth}
        \centering
        \includegraphics[width=\textwidth]{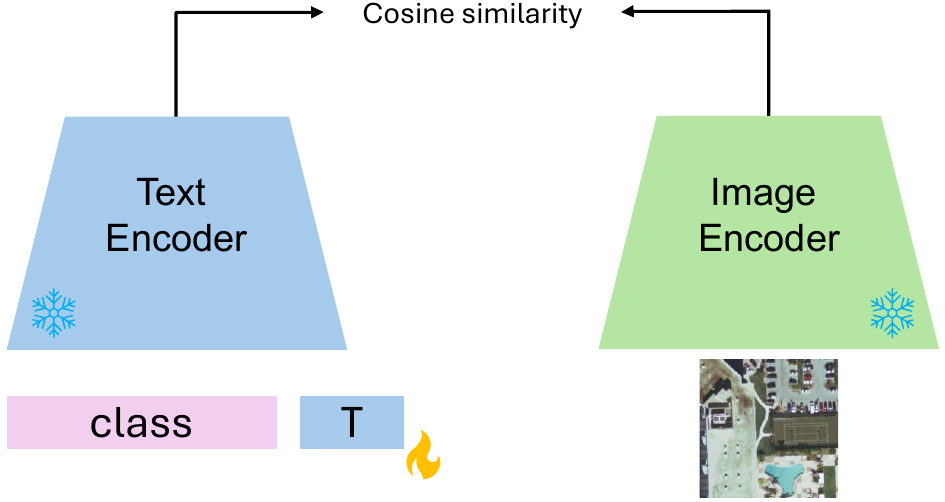}
        \caption{CoOp}
        \label{fig:coop}
    \end{subfigure}
    \hfill
    \begin{subfigure}[t]{0.45\textwidth}
        \centering
        \includegraphics[width=\textwidth]{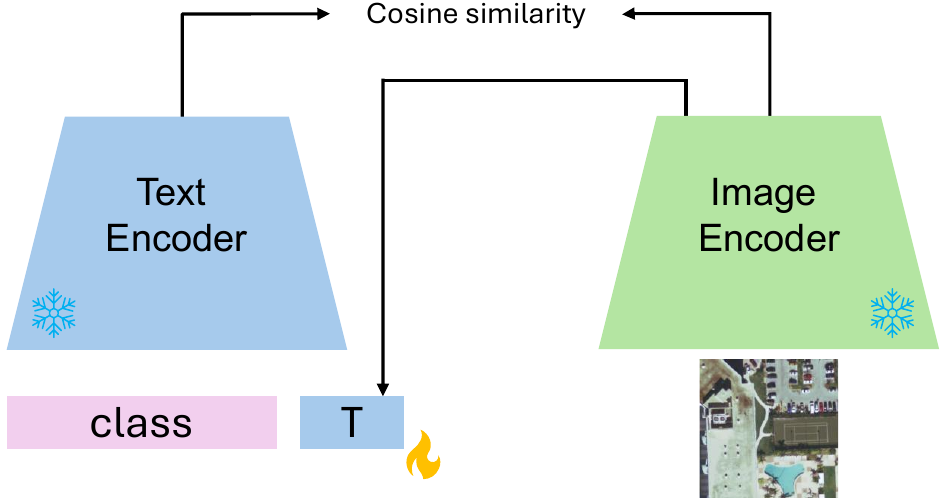}
        \caption{CoCoOp}
        \label{fig:cocoop}
    \end{subfigure}

    \vspace{1em} % space between rows

    % ---------- Row 2 ----------
    \begin{subfigure}[t]{0.45\textwidth}
        \centering
        \includegraphics[width=\textwidth]{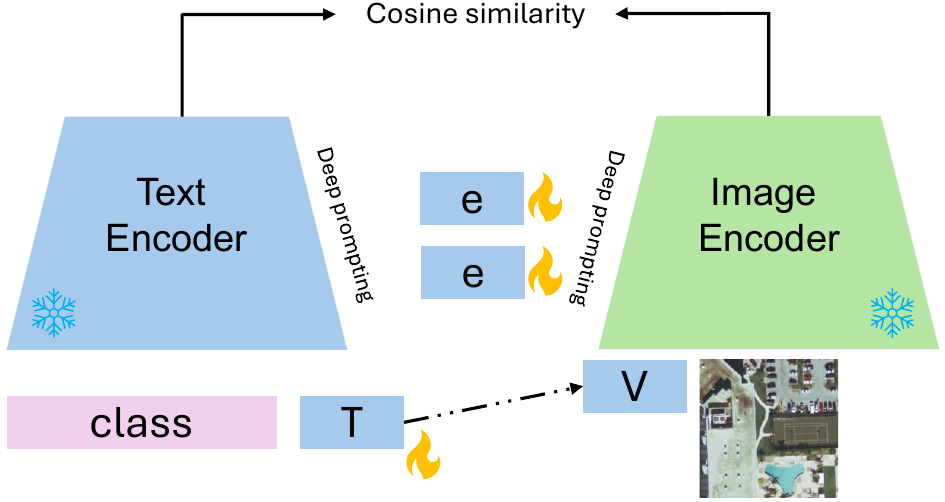}
        \caption{MaPLe}
        \label{fig:maple}
    \end{subfigure}
    \hfill
    \begin{subfigure}[t]{0.45\textwidth}
        \centering
        \includegraphics[width=\textwidth]{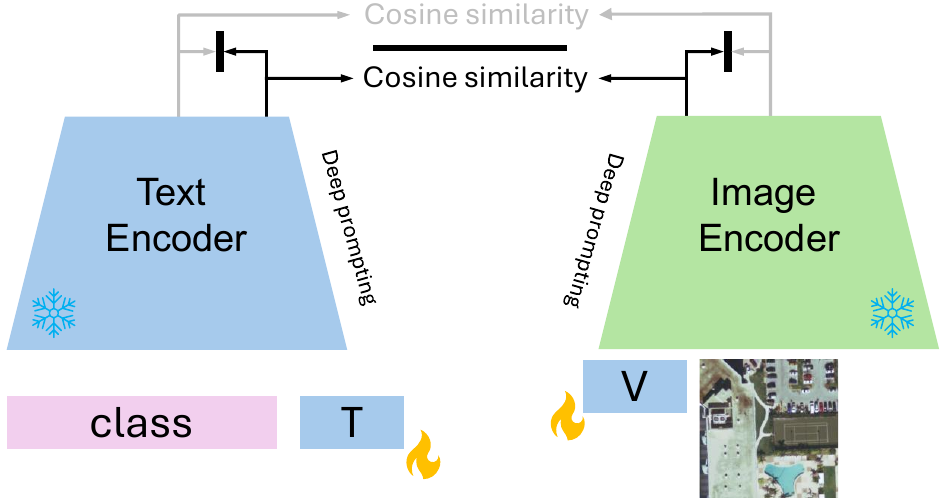}
        \caption{PromptSRC}
        \label{fig:promptscr}
    \end{subfigure}

    \caption{Comparative illustration of the four prompt-learning paradigms evaluated in this study. In the figure, the snowflake icons denote frozen parameters, while the flame icons indicate learnable components. Deep prompting refers to learnable tokens inserted across multiple layers of the encoder. The symbols T and V represent learnable text and visual prompt embeddings, respectively. Cosine similarity denotes the similarity computation used for fine-tuning, whereas the gray cosine similarity indicates the same computation within the frozen architecture. In the MaPLe diagram, the symbol "e" denotes the coupling function that links the text and vision encoders across several layers.
    (a)~CoOp learns static textual context embeddings appended to class name tokens.
    (b)~CoCoOp introduces instance-conditioned prompts for enhanced generalization.
    (c)~MaPLe jointly learns textual and visual prompts for cross-modal coupling.
    (d)~PromptSRC employs self-regulating constraints to preserve pre-trained alignment while adapting to new tasks.}
    \label{fig:prompt-learning}
\end{figure*}

PromptSRC extends prompt learning by introducing a self-regulating mechanism that enables adaptation without catastrophic forgetting of the pre-trained knowledge~\cite{khattak2023self}. Building upon CLIP’s dual-encoder architecture, PromptSRC inserts learnable prompt tokens into both the vision and text branches but constrains their optimization through a mutual-agreement objective that aligns the prompted and frozen feature spaces. The overall PromptSRC framework, highlighting its self-regulating alignment between prompted and unprompted representations, is depicted in Figure~\ref{fig:promptscr}.

The key idea is to preserve the generalization ability of the pre-trained CLIP model while fine-tuning it for specific downstream tasks. To achieve this, the model produces two parallel representations using the same frozen encoders: one derived from inputs augmented with learnable prompts, and another obtained without prompts, representing the original pre-trained behavior. A self-regulating constraint is then applied to maximize agreement between these two feature spaces at both the feature and logit levels. This mechanism ensures that task-specific adaptation does not distort the intrinsic vision-language alignment learned during large-scale pre-training.

In addition to this self-regulating objective, PromptSRC introduces textual diversity by generating multiple augmented versions of the text prompts, which encourages the model to remain semantically robust across phrasing variations. During training, prompt parameters from different epochs are further combined through a self-ensembling strategy, yielding more stable convergence and improved generalization. The overall effect is a model that adapts effectively to new domains while retaining the broad semantic understanding of its foundation model.

In remote sensing applications, where datasets are often limited and exhibit substantial domain shifts, PromptSRC offers a strong balance between adaptability and stability. By regulating prompt updates relative to CLIP’s original representations, it mitigates overfitting and maintains alignment across spectral, spatial, and temporal variations, making it particularly suited for few-shot and cross-dataset scene classification tasks.

%%%%%%%%%%%%%%%%%%%%%%%%%%%%%%%%%%%%%%%%%%
\section{Experimental design}
This section describes the experimental setup used to evaluate the effectiveness of prompt-learning methods to adapt CLIP to remote sensing image scene classification. We describe datasets, model architecture, baseline methods, few-shot training configurations, implementation setup, and evaluation metrics. The objective is to establish a transparent and reproducible framework for fair comparison across zero-shot, linear probing, and multiple prompt-learning approaches.

\subsection{Datasets and pre-processing}
We handle multiple remote sensing datasets using a modular framework. Each dataset is encapsulated in a dedicated class that provides a unified interface for accessing images, class labels, and annotations. The dataset splits follow the AITLAS Arena protocol \cite{rs15092343, DIMITROVSKI202318}, ensuring consistent and reproducible evaluation. Pre-defined test splits are used across all models to guarantee fair comparisons.

Raw dataset labels in remote sensing collections often contain underscores, technical abbreviations, or pluralized forms that are not suitable for natural-language prompts. To ensure linguistic consistency, we normalize class names as follows: (i) underscores are replaced with spaces (e.g., mobile\_home\_park is replaced with “mobile home park”); (ii) uncommon or compound terms are rephrased into standard English (e.g., bareland is replaced with “bare land”); and (iii) class forms are standardized for clarity (e.g., AnnualCrop is replaced with “annual crop land”). The complete mapping from raw to normalized labels is available in our public repository for full reproducibility.

\subsection{Model architecture and baselines}
All experiments are carried out using the CLIP model with a ViT-B/16 Transformer architecture \cite{dosovitskiy2020image, radford2021learning}. The models are implemented on top of OpenCLIP \cite{cherti2023reproducible}, the open source implementation of OpenAI's CLIP. No architectural modifications are made to CLIP, only the prompt parameters are trainable.

We benchmark the four prompt-learning approaches CoOp, CoCoOp, MaPLe, and PromptSRC against two strong baselines. The first baseline is zero-shot CLIP, which employs the frozen CLIP model with manually crafted textual prompts following \cite{radford2021learning}. For remote sensing imagery, we use the template "a satellite photo of \{class\}" to ensure semantic alignment with the visual domain.

The second is the linear probe approach, which trains a linear classifier on top of frozen CLIP image embeddings, following the procedure outlined in \cite{radford2021learning, tian2020rethinking}. We freeze the CLIP visual encoder and train an L2-regularized multinomial logistic regression classifier on the extracted features. We use the lbfgs solver\footnote{https://scikit-learn.org/stable/modules/generated/sklearn.linear\_model.LogisticRegression.html} with a maximum of 1000 iterations. The regularization strength C is chosen via a two-stage procedure: first, a coarse grid search and then, up to 8 refinement steps of binary search to narrow the optimal range. We report results averaged over 3 random seeds, each with independent few-shot sampling. No additional augmentations are applied beyond the standard CLIP pre-processing pipeline.

\subsection{Few-shot evaluation protocol}
\label{section:few_shot_setup}
For a few-shot evaluation, we follow the protocol in \cite{radford2021learning}, assessing the model's ability to adapt to new tasks with a very limited number of training examples. The evaluation was conducted using varying numbers of "shots" per class, specifically 1, 2, 4, 8, and 16 images per class from the official training split of each dataset. This range of shot counts allows for a comprehensive analysis of how each model's performance scales with the amount of available training data, from extremely limited to moderately low. Class balance was strictly enforced: for each class, exactly $k$ images were drawn, ensuring equal representation across all classes. To reduce variance, we repeated this sampling procedure across three independent runs. The random seed controlling the sampling was fixed per run, and the same sampled subsets were used consistently across all methods (linear probe, CoOp, CoCoOp, MaPLe, and PromptSRC) to guarantee fairness in comparison. For each shot setting, models are evaluated on the full predefined test sets, and the reported results are averaged over these three independent runs.

\subsection{Cross-dataset evaluation}
To evaluate the generalization ability of prompt-learning methods beyond their training domains, we perform a cross-dataset evaluation. In this setup, we reuse the models trained under the 16-shot configuration from the few-shot experiments (Section~\ref{section:few_shot_setup}) and directly test them on unseen datasets without any additional fine-tuning or adaptation. This procedure assesses how well the learned prompts transfer across different remote sensing domains that vary in spatial resolution, acquisition conditions, and scene semantics.

All prompt-learning models CoOp, CoCoOp, MaPLe, and PromptSRC are included in this evaluation. For each experiment, a model trained on a given \textit{source} dataset is tested on the full test split of every other \textit{target} dataset, producing a complete source-target performance matrix. This cross-dataset setup enables systematic analysis of domain transferability and allows us to identify which prompting strategies exhibit stronger generalization behavior.

To ensure fairness and reproducibility, all evaluations use identical model checkpoints obtained from the 16-shot training regime, along with consistent pre-processing, random seeds, and optimization parameters. Results are reported in terms of overall accuracy averaged across three runs, and qualitative transfer patterns are further analyzed using cross-domain confusion matrices.

\subsection{Training and implementation details}
Training and evaluation were performed on a single NVIDIA GeForce GTX 1080 Ti GPU with 11 GB of memory, running CUDA version 12.2. CoCo, CoCoOp, MaPLe, and PromptSRC follow their respective published configurations \cite{zhou2022learning}, \cite{zhou2022conditional}, \cite{khattak2023maple}, \cite{khattak2023self}, with hyper-parameters aligned for consistency across experiments to ensure comparability under identical training regimes.

For CoOp and CoCoOp, we adopt the end-position class token design in which the class token is placed at the end of the sequence. we fix the
context length to 4 and initialize the context vectors using the pre-trained word embeddings of “a photo of a” for both CoOp and CoCoOp. Training is performed using Stochastic Gradient Descent (SGD) with an initial learning rate of 0.002, decayed using a cosine annealing schedule. For MaPLe, we set prompt depth to 9 and the language and vision prompt lengths to 2. Training is performed with a learning rate of 0.0035 via SGD optimizer. We initialize the language prompts of the first layer with the pretrained CLIP word embeddings of the template "a photo of a", while for the subsequent layers they are randomly initialized from a normal distribution.

For PromptSRC, we adopt a deep prompting strategy with 4 vision prompt tokens and 4 language prompt tokens. Prompts are inserted and optimized across the first nine transformer layers of the CLIP backbone. All prompts are randomly initialized from a normal distribution, except for the text prompts in the first layer, which are initialized using the word embeddings of the template "a photo of a". The learning rate is fixed to 0.0025 for all runs. The weighting coefficients for the self-consistency losses are set to $\lambda_1 = 10$ and $\lambda_2 = 25$ for the image and text branches, respectively. These hyper-parameters remain constant across all datasets and benchmarks to ensure fair and consistent comparisons. For textual diversity, we employ $N = 60$ standard prompt templates as introduced in \cite{khattak2023self}, allowing for greater linguistic variation during training.

All models are trained for 50 epochs with a batch-size of 4. To stabilize early training, a learning rate warm-up is applied by fixing the rate to $1 \times 10^{-5}$ during the first epoch. 

\subsection{Evaluation metrics}
We report top-1 overall accuracy as the main quantitative metric and additionally visualize normalized confusion matrices for qualitative per-class error analysis. 

\section{Results and discussion}
\begin{figure}
\centering
  \includegraphics[trim={0 0 0 0},clip, width=1.0\linewidth]{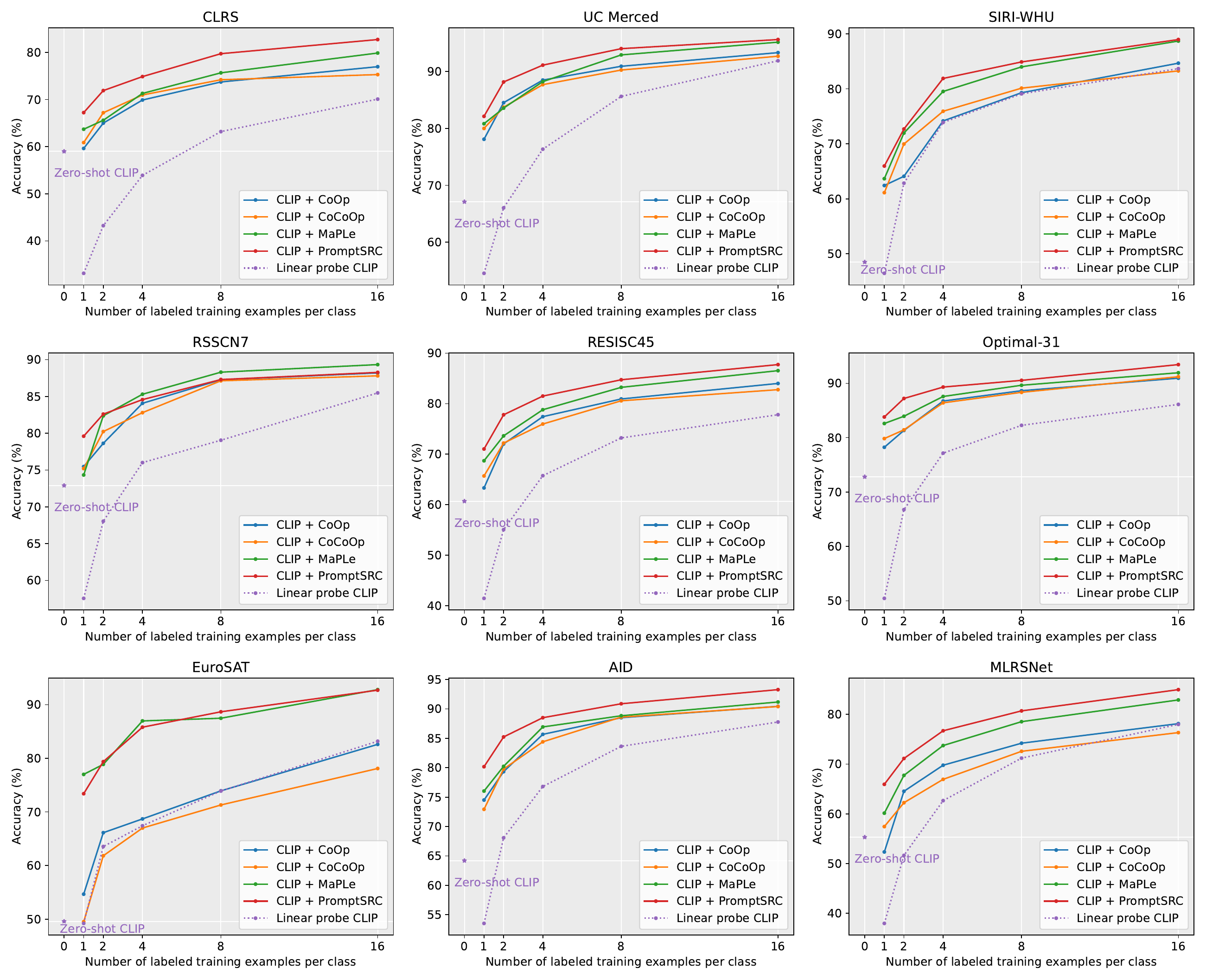}
  \caption{Zero-shot and few-shot classification results across nine remote sensing datasets. 
Each plot shows performance as a function of the number of shots for four prompt-learning methods (CoOp, CoCoOp, MaPLe, and PromptSRC) compared with zero-shot CLIP (hand-crafted prompts) and the linear probe baseline. Prompt-learning approaches consistently outperform both baselines, with MaPLe and PromptSRC achieving the strongest gains in few-shot scenarios.}
  \label{fig:results_datasets}
\end{figure}

\begin{figure}
\centering
  \includegraphics[trim={0 0 0 0},clip, width=0.6\linewidth]{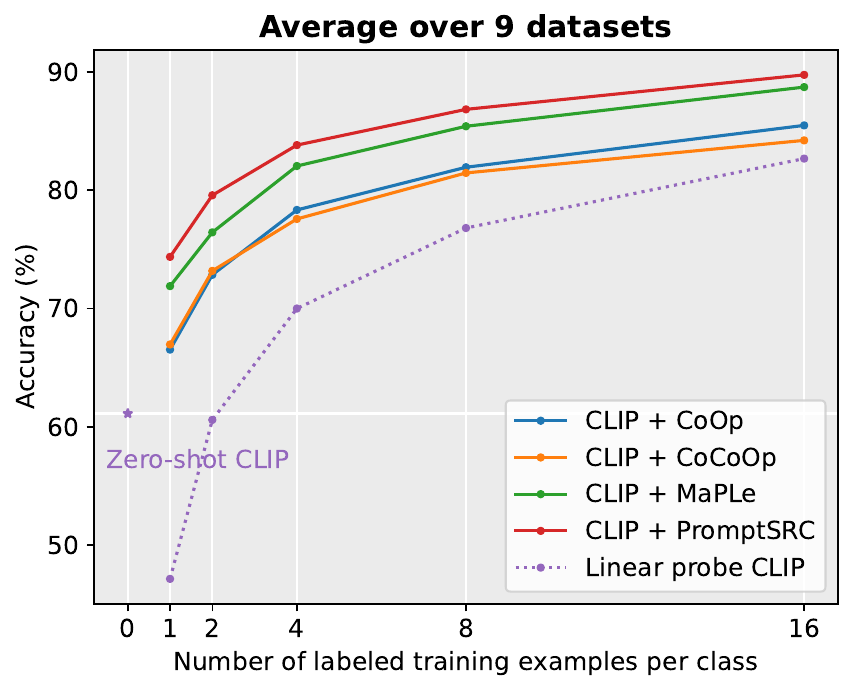}
  \caption{Average zero-shot and few-shot classification accuracy across the nine benchmark remote sensing datasets. Results compare zero-shot CLIP (hand-crafted prompts), linear probe, and four prompt-learning approaches (CoOp, CoCoOp, MaPLe, and PromptSRC). The plot highlights the consistent improvement achieved by prompt-learning methods, with PromptSRC showing the highest overall accuracy and MaPLe achieving strong cross-modal generalization.}
  \label{fig:average_results}
\end{figure}

Table~\ref{tab:fewshot_results} reports the zero-shot and few-shot classification accuracy across nine benchmark remote sensing datasets, while Figure~\ref{fig:results_datasets} provides a complementary visualization of the same results by plotting performance as a function of the number of shots for each method. This dual presentation allows both a detailed inspection of the numerical results and a clearer illustration of the relative trends across methods and shot settings. In addition, Figure~\ref{fig:average_results} summarizes the average performance across all datasets, offering a concise comparison of the overall trends. Several consistent patterns emerge from the experiments. First, zero-shot CLIP with hand-crafted prompts provides a strong baseline, achieving accuracies between 48.50\% and 72.90\%, depending on the dataset. However, the performance gap between zero-shot and few-shot learning is substantial, confirming that handcrafted textual templates alone are insufficient for specialized domains such as remote sensing.

\begin{table*}
\centering
\caption{Zero-shot and few-shot classification accuracy (\%) of different methods across nine remote sensing datasets. For each dataset, the best-performing method is highlighted in bold and the second best is underlined.}
\resizebox{\textwidth}{!}{
%\begin{adjustbox}{angle=90,max totalsize={\textheight}{\textwidth},center}
\begin{tabular}{lccccccccc}
\toprule
\textbf{Method} & \textbf{EuroSAT} & \textbf{UC Merced} & \textbf{RESISC45} & \textbf{AID} & \textbf{RSSCN7} & \textbf{Optimal-31} & \textbf{SIRI-WHU} & \textbf{CLRS} & \textbf{MLRSNet} \\
\midrule
Handcrafted prompt & 49.60 & 67.10 & 60.70 & 64.20 & 72.90 & 72.80 & 48.50 & 59.00 & 55.30 \\
\midrule
CLIP+CoOp, shots=1 & 54.67 & 78.10 & 63.33 & 74.50 & 75.47 & 78.23 & 62.43 & 59.63 & 52.33 \\
CLIP+CoOp, shots=2 & 66.13 & 84.53 & 72.03 & 79.33 & 78.63 & 81.33 & 64.10 & 65.00 & 64.53 \\
CLIP+CoOp, shots=4 & 68.70 & 88.50 & 77.43 & 85.67 & 84.07 & 86.73 & 74.17 & 69.90 & 69.77 \\
CLIP+CoOp, shots=8 & 73.93 & 90.93 & 80.93 & 88.50 & 87.27 & 88.63 & 79.30 & 73.73 & 74.20 \\
CLIP+CoOp, shots=16 & 82.60 & 93.33 & 84.00 & 90.43 & 88.20 & 90.93 & 84.67 & 76.97 & 78.13 \\
\midrule
CLIP+CoCoOp, shots=1 & 49.53 & 80.00 & 65.70 & 72.93 & 75.20 & 79.83 & 61.13 & 60.87 & 57.43 \\
CLIP+CoCoOp, shots=2 & 61.83 & 83.73 & 72.20 & 79.77 & 80.23 & 81.43 & 69.97 & 67.20 & 62.23 \\
CLIP+CoCoOp, shots=4 & 67.00 & 87.70 & 75.97 & 84.40 & 82.80 & 86.40 & 75.93 & 70.97 & 66.93 \\
CLIP+CoCoOp, shots=8 & 71.30 & 90.27 & 80.57 & 88.60 & 87.13 & 88.33 & 80.13 & 74.20 & 72.57 \\
CLIP+CoCoOp, shots=16 & 78.10 & 92.70 & 82.77 & 90.40 & 87.80 & 91.20 & 83.27 & 75.30 & 76.33 \\
\midrule
CLIP+MaPLe, shots=1 & 77.00 & 80.83 & 68.70 & 76.03 & 74.33 & 82.60 & 63.67 & 63.70 & 60.13 \\
CLIP+MaPLe, shots=2 & 78.87 & 83.57 & 73.63 & 80.23 & 82.37 & 83.93 & 71.97 & 65.60 & 67.73 \\
CLIP+MaPLe, shots=4 & 86.97 & 88.23 & 78.80 & 86.93 & 85.30 & 87.57 & 79.53 & 71.30 & 73.73 \\
CLIP+MaPLe, shots=8 & 87.47 & 92.93 & 83.23 & 88.83 &  \underline{88.30} & 89.63 & 84.00 & 75.67 & 78.53 \\
CLIP+MaPLe, shots=16 & \textbf{92.80} &  \underline{95.17} &  \underline{86.53} &  \underline{91.17} & \textbf{89.33} &  \underline{91.93} &  \underline{88.70} &  \underline{79.87} &  \underline{82.90} \\
\midrule
CLIP+PromptSRC, shots=1 & 73.40 & 82.13 & 71.03 & 80.17 & 79.60 & 83.80 & 65.97 & 67.23 & 65.93 \\
CLIP+PromptSRC, shots=2 & 79.37 & 88.17 & 77.80 & 85.23 & 82.60 & 87.20 & 72.70 & 71.90 & 71.13 \\
CLIP+PromptSRC, shots=4 & 85.80 & 91.13 & 81.50 & 88.50 & 84.57 & 89.30 & 81.90 & 74.87 & 76.70 \\
CLIP+PromptSRC, shots=8 & 88.67 & 94.03 & 84.73 & 90.87 & 87.30 & 90.53 & 84.90 & 79.73 & 80.70 \\
CLIP+PromptSRC, shots=16 & \underline{92.70} & \textbf{95.63} & \textbf{87.73} & \textbf{93.27} & 88.27 & \textbf{93.43} & \textbf{88.97} & \textbf{82.73} & \textbf{84.97} \\
\midrule
linear probe CLIP, shots=1 & 49.22 & 54.52 & 41.48 & 53.53 & 57.56 & 50.45 & 46.46 & 33.13 & 37.96 \\
linear probe CLIP, shots=2 & 63.55 & 66.03 & 55.07 & 68.08 & 68.04 & 66.76 & 62.85 & 43.24 & 51.63 \\
linear probe CLIP, shots=4 & 67.42 & 76.35 & 65.75 & 76.80 & 76.01 & 77.15 & 73.89 & 53.91 & 62.65 \\
linear probe CLIP, shots=8 & 73.91 & 85.63 & 73.22 & 83.63 & 79.05 & 82.26 & 79.10 & 63.20 & 71.20 \\
linear probe CLIP, shots=16 & 83.18 & 91.90 & 77.82 & 87.77 & 85.48 & 86.11 & 83.68 & 70.09 & 77.98 \\
\bottomrule
\end{tabular}
}
%\end{adjustbox}
\label{tab:fewshot_results}
\end{table*}

Second, both CoOp and the linear probe baselines show a clear trend of improvement as the number of training shots increases. Even in the extremely low-shot regime (1-2 samples per class), CoOp achieves notable gains over zero-shot CLIP, confirming its effectiveness as a lightweight adaptation mechanism. Linear probing remains less effective with limited data but becomes increasingly competitive at higher shot counts, benefiting from more supervision to tune the shallow classifier.

The conditional extension, CoCoOp, yields marginal improvements over CoOp in low-shot regimes and on datasets with large intra-class variability. The adaptive Meta-Net allows prompts to better capture instance-level cues, improving robustness when the training distribution is limited. However, as the number of shots increases, CoOp and CoCoOp converge to similar performance, suggesting that conditional adaptation provides the largest benefit in data-scarce scenarios.

Multi-modal Prompt Learning (MaPLe) consistently outperforms both CoOp and CoCoOp across most benchmarks. By jointly optimizing prompts in both the vision and language branches and introducing cross-modal coupling, MaPLe achieves stronger alignment between modalities. This leads to particularly high accuracies on complex datasets with greater spectral and semantic diversity, surpassing $92\%$ on EuroSAT and over $95\%$ on UC Merced. The results confirm that deep, coupled prompting enhances generalization to fine-grained remote sensing categories.

The best overall results are obtained with PromptSRC, which integrates self-regulating constraints to stabilize learning and prevent overfitting. Its performance is consistently the highest or second highest across all datasets, reaching up to $93.4\%$ on Optimal-31 and $85.0\%$ on MLRSNet. These improvements demonstrate that incorporating self-consistency and diversity regularization helps maintain generalizable representations even with limited training samples.

When compared to the linear probe, all prompt-learning approaches achieve higher accuracy, particularly in the low-shot regime. This confirms that learnable prompts adapt the pre-trained vision-language representations more effectively than training a shallow classifier on frozen embeddings. The gap narrows at higher shot counts, but prompt-learning methods retain a clear advantage in data efficiency and transferability.

To complement the overall accuracy results, we further examined the per-class performance of the best-performing models (typically PromptSRC, and in a few cases MaPLe) by computing normalized confusion matrices for all nine datasets. These analyses provide a detailed view of recognition strengths and systematic misclassification patterns.

Across datasets, the confusion matrices reveal that the model consistently achieves high per-class accuracy for visually distinctive categories, such as airports, runways, beaches, mountains, and water bodies, with many diagonal entries exceeding 0.90. However, systematic confusions emerge in semantically or structurally related categories. Examples include mutual confusion among residential classes (dense, medium, sparse), difficulties distinguishing railway from railway station, and overlaps between vegetation types such as crops, pastures, and meadows. Water-related categories (river, pond, lake) also show misclassifications in several datasets.  

Overall, the results demonstrate a clear performance hierarchy among prompt-learning methods: CoOp and CoCoOp provide effective low-data adaptation, MaPLe enhances cross-modal representation learning, and PromptSRC achieves the strongest generalization through self-regularization. Together, these findings highlight prompt learning as a powerful and scalable strategy for adapting vision-language models to remote sensing imagery, especially under limited data availability. Challenges remain in fine-grained discrimination of visually or semantically adjacent categories. Potential improvements may come from refined prompt designs, incorporation of spatial-contextual priors, or hierarchical classification strategies.

For completeness, the normalized confusion matrices with detailed per-dataset discussions are provided in Appendix~\ref{appendix:confusion_matrices}.

\subsection{Cross-dataset generalization analysis}
\begin{figure}[H]
\centering
  \includegraphics[trim={0 0 0 0},clip, width=0.9\linewidth]{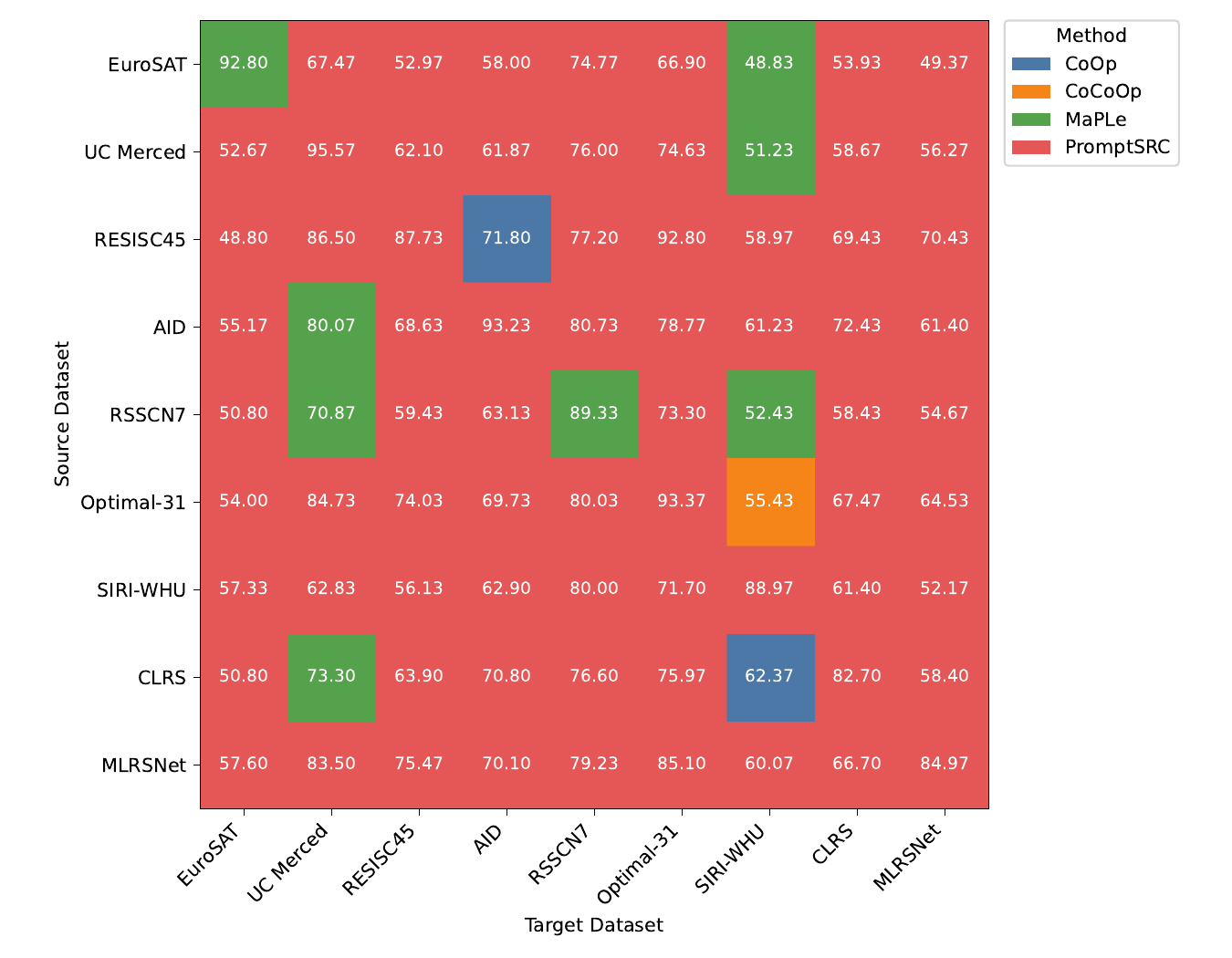}
  \caption{Element-wise comparison of cross-dataset classification accuracies among four prompt-learning methods (CoOp, CoCoOp, MaPLe, and PromptSRC). Each cell represents the maximum accuracy obtained across the four methods for a given source-target dataset pair, with color coding indicating the method that achieved the highest performance.}
  \label{fig:cross_dataset_results}
\end{figure}

To assess the generalization ability of the evaluated prompt-learning approaches, we conducted a cross-dataset transfer experiment in which each model trained on a source dataset was tested on all other target datasets. The resulting winner matrix presented in Figure~\ref{fig:cross_dataset_results} reveals that PromptSRC dominates most source-target combinations, highlighting its robustness to domain shifts and diverse scene distributions. MaPLe demonstrates competitive performance in several challenging cross-domain pairs, particularly those involving UC Merced and RSSCN7, indicating that its multi-modal context learning effectively adapts to varying feature distributions. In contrast, CoOp and CoCoOp outperform the others only in isolated cases (e.g., RESISC45 → AID and Optimal-31 → SIRI-WHU), where the visual or semantic domains of the datasets are closely related.

The maximum-value matrix shows that cross-dataset accuracy generally drops compared to the diagonal self-evaluation results, reflecting the inherent difficulty of transferring knowledge between remote sensing datasets with different resolutions, acquisition conditions, and label semantics. Nonetheless, the relatively high off-diagonal values obtained by PromptSRC and MaPLe suggest improved transferability of their learned prompt representations. Overall, the analysis demonstrates that PromptSRC achieves the most consistent cross-dataset generalization, while MaPLe remains a strong alternative for heterogeneous domain adaptation scenarios.

%%%%%%%%%%%%%%%%%%%%%%%%%%%%%%%%%%%%%%%%%%
\section{Conclusions}
This paper investigated prompt learning as a lightweight and effective strategy for adapting vision-language foundation models to remote sensing image scene classification. Building on CLIP with a ViT-B/16 backbone, we systematically evaluated four prompt-learning paradigms: Context Optimization (CoOp), Conditional Context Optimization (CoCoOp), Multi-modal Prompt Learning (MaPLe), and Prompting with Self-regulating Constraints (PromptSRC), alongside two representative baselines: zero-shot CLIP with handcrafted textual prompts and a linear probe on frozen image embeddings.

Across nine benchmark datasets and multiple few-shot settings, the results demonstrate that prompt learning significantly enhances CLIP’s adaptability to remote sensing imagery without modifying the pre-trained architecture or requiring large-scale domain-specific pretraining. CoOp and CoCoOp deliver strong improvements in low-data regimes, validating the role of learnable context tokens for data-efficient adaptation. MaPLe further strengthens performance through joint vision-language prompting and cross-modal coupling, while PromptSRC achieves the highest overall accuracy and stability by integrating self-regulating constraints that mitigate overfitting and promote generalization.

An analysis of confusion matrices for the best-performing configurations reveals that prompt-based models reliably recognize visually distinctive land-cover categories, whereas residual errors remain concentrated among semantically similar or spectrally overlapping classes. These challenges point toward future research directions such as incorporating spatial-contextual priors, hierarchical category structures, and multi-temporal representations to improve fine-grained scene discrimination.

Overall, the findings highlight prompt learning as a scalable, data-efficient, and architecture-agnostic pathway for adapting foundation vision-language models to remote sensing tasks. As prompt-based methods continue to evolve, they offer a promising foundation for building next-generation Earth observation models that combine strong generalization with minimal supervision.

Building upon the findings of this study, several promising directions emerge for future research. First, fine-grained scene understanding could benefit from hierarchical and spatial-contextual prompting strategies that capture local structures and semantic relationships within complex urban and agricultural environments. Second, integrating multi-temporal and multi-modal data sources, including multispectral, hyperspectral, and SAR imagery could further enhance the generalization capacity of prompt-learning frameworks across sensors and observation conditions. Third, exploring cross-domain and cross-sensor transfer would enable broader applicability of learned prompts beyond dataset-specific distributions. Finally, optimizing prompt-tuning for computational efficiency and deploying lightweight versions on edge devices such as UAVs and satellite platforms represent important steps toward operational remote sensing applications. Together, these directions outline a path toward more adaptive, interpretable, and scalable vision-language systems for Earth observation.


Author Contributions I.D.: Methodology, Investigation, Software, Writing, Visualization V.S.: Methodology, Writing, Visualization I.K.: Methodology, Writing, Visualization All authors reviewed and agreed to the final version of the manuscript.

Data Availability The code to reproduce the results from the paper will be available at https://github.com/ivicadimitrovski upon submission decision.

Funding The authors declare that no funding was received for conducting this study.

\begin{appendices}
\section{Confusion Matrices for Each Dataset}
\label{appendix:confusion_matrices}

This appendix presents detailed normalized confusion matrices for each dataset, along with per-class performance analysis and discussion of systematic misclassifications. All results correspond to the best-performing configuration on the specific dataset.

\subsection{EuroSAT}
\begin{figure}[H]
\centering
  \includegraphics[trim={0 0 0 0},clip, width=0.7\linewidth]{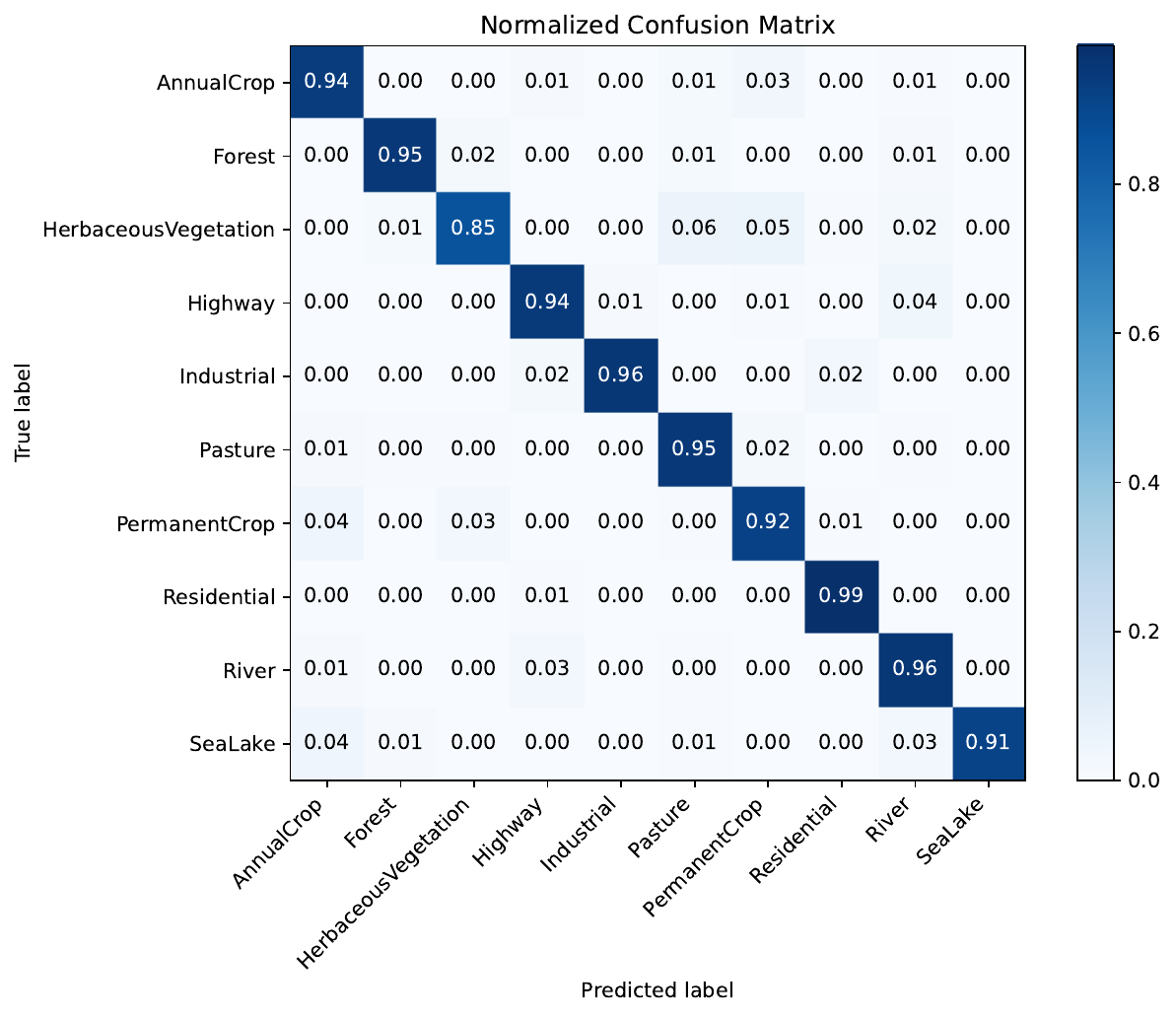}
  \caption{Normalized confusion matrix illustrating per-class performance of the best-performing configuration (CLIP-MaPLe, 16-shot setting) on the EuroSAT dataset.}
  \label{fig:cmat_eurosat}
\end{figure}

The normalized confusion matrix for EuroSAT, presented in Figure~\ref{fig:cmat_eurosat} reveals strong performance across most land-cover classes, with diagonal entries generally exceeding 0.85, indicating high per-class accuracy. Classes such as Residential, River, and Industrial are classified with the highest reliability, reflecting their distinctive spectral and spatial signatures. However, systematic misclassifications are also evident. For instance, AnnualCrop is occasionally confused with PermanentCrop and Pasture, highlighting the difficulty in distinguishing vegetation types with similar textures and seasonal variability. Likewise, Herbaceous Vegetation shows overlap with Pasture and PermanentCrop, suggesting that shared vegetation characteristics can blur class boundaries. In addition, River exhibits confusion with Highway and AnnualCrop, which likely stems from the presence of linear features or mixed land-cover contexts near waterways.

Overall, the EuroSAT results demonstrate that the model is well-suited for broad land-cover discrimination but faces challenges in fine-grained vegetation categories and classes with strong contextual adjacency effects. These patterns suggest that further improvements may be achieved by refining prompt engineering strategies or incorporating auxiliary spatial-contextual cues to better disambiguate visually similar land-cover types.

\subsection{UC Merced}
\begin{figure}[H]
\centering
  \includegraphics[trim={0 0 0 0},clip, width=1.0\linewidth]{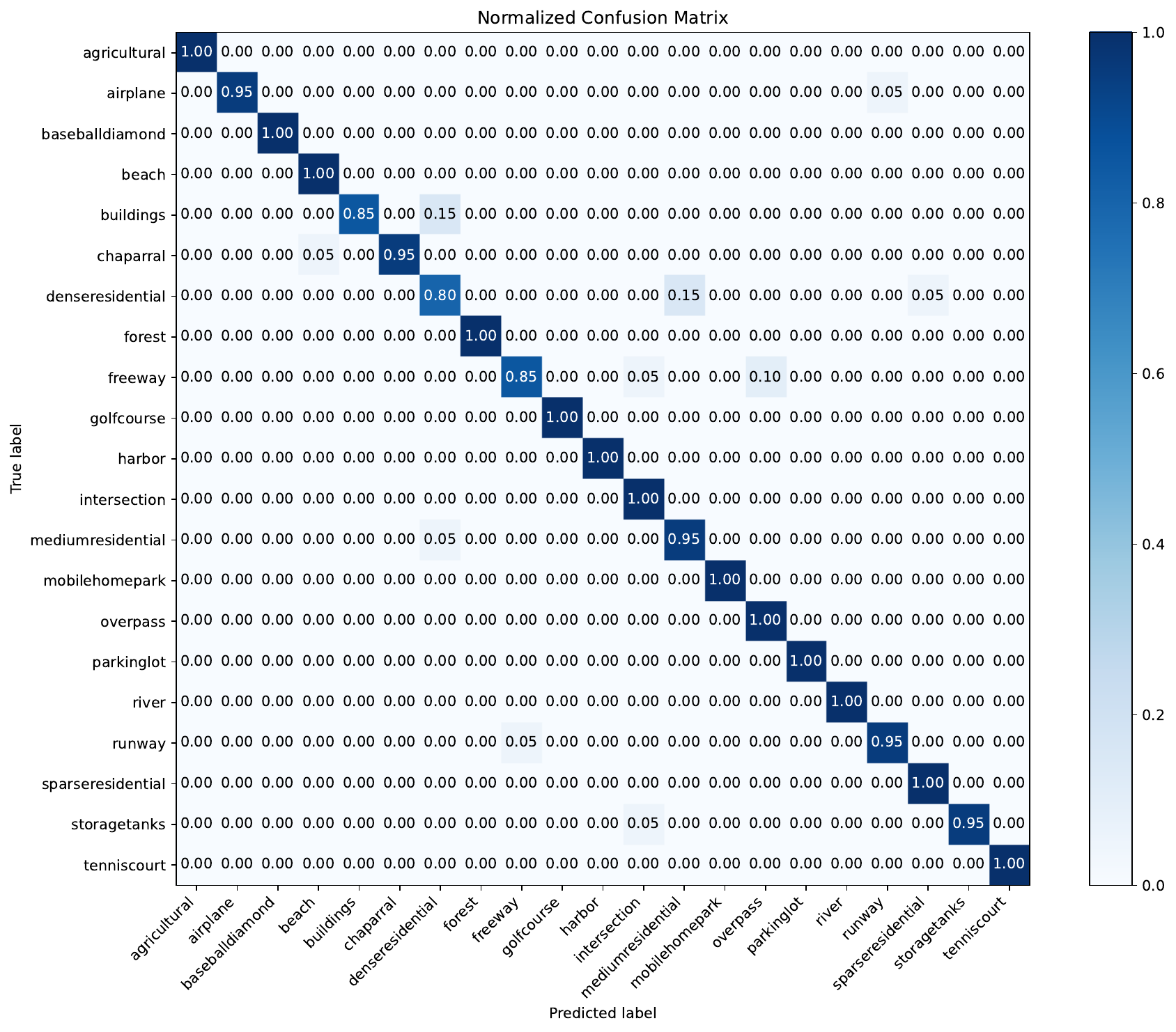}
  \caption{Normalized confusion matrix illustrating per-class performance of the best-performing configuration (CLIP-PromptSRC, 16-shot setting) on the UC Merced dataset.}
  \label{fig:cmat_ucmerced}
\end{figure}

The normalized confusion matrix for UC Merced, presented in Figure~\ref{fig:cmat_ucmerced} demonstrates exceptionally strong performance across all 21 scene categories, with nearly all diagonal values approaching 1.0. This indicates that the model achieves near-perfect discrimination between the majority of classes in this dataset. High recognition accuracy is observed for visually distinctive categories such as agricultural, baseballdiamond, beach, forest, golf course, harbor, runway, storage tanks, and tennis court, which show no or very limited misclassification.

Despite the overall robustness, certain systematic confusions can be identified in the off-diagonal entries. The buildings are sometimes misclassified as dense residential, reflecting the challenge of separating generic building patterns from high-density urban areas. The dense residential class itself shows misclassifications into medium residential and sparse residential, which is expected given the shared urban structures and gradual transitions in housing density. The class airplane also exhibits a small degree of confusion with the class runway, and the class freeway with the classes intersection and overpass suggesting structural similarities in aerial layouts.

Overall, the UC Merced results highlight the model’s excellent capacity to capture discriminative spatial-structural features of overhead imagery, while residual confusions are largely restricted to semantically or visually related urban categories. This outcome confirms that foundation models adapted with prompt learning are particularly effective on datasets with well-defined scene semantics, such as UC Merced, but that fine-grained separation of related residential or infrastructural classes remains a challenge.

\subsection{RESISC45}
\begin{figure}[H]
\centering
  \includegraphics[trim={0 0 0 0},clip, width=1.0\linewidth]{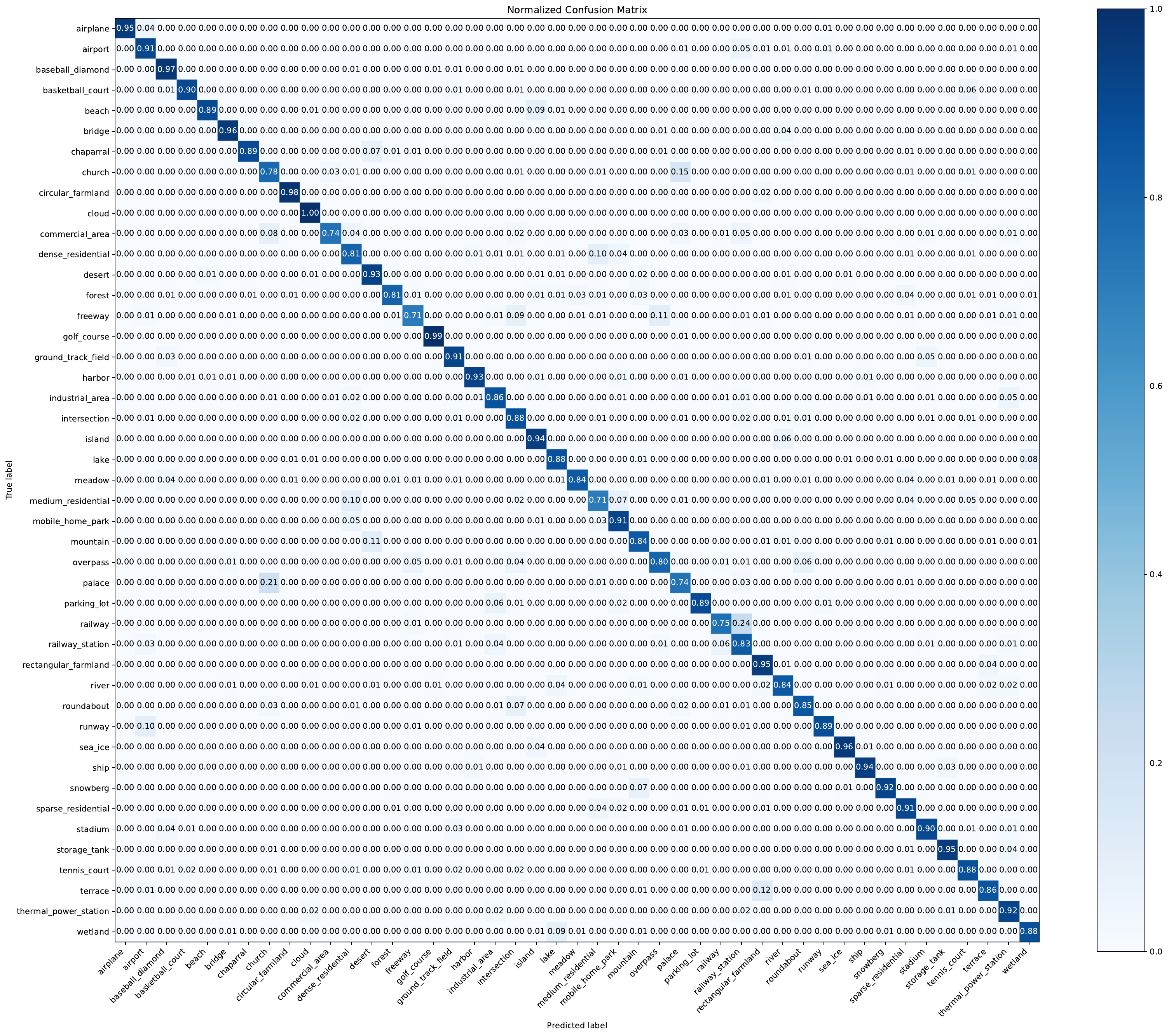}
  \caption{Normalized confusion matrix illustrating per-class performance of the best-performing configuration (CLIP-PromptSRC, 16-shot setting) on the RESISC45 dataset.}
  \label{fig:cmat_resisc45}
\end{figure}

The normalized confusion matrix for RESISC45, presented in Figure~\ref{fig:cmat_resisc45} indicates strong overall performance across the 45 scene categories, with most diagonal entries above 0.85. Distinctive classes such as airplane, baseball diamond, bridge, circular farmland, cloud, golf course, rectangular farmland, sea ice, and storage tank are classified with very high reliability, with per-class accuracies often exceeding 0.95. These results highlight the model’s ability to capture distinctive structural and spectral patterns associated with highly separable land-use types.

However, the off-diagonal elements reveal notable challenges in certain visually or semantically related categories. Dense residential, medium residential, and sparse residential show mutual confusion, reflecting gradual transitions in urban density and shared textural cues across housing areas. Similarly, railway and railway station exhibit substantial confusion, which is expected given their semantic and spatial proximity in aerial imagery. Commercial area is occasionally misclassified as industrial area, likely due to overlapping built-environment structures. Confusion between river and lake further illustrates the difficulty in disambiguating classes that share similar vegetation or water features.

Overall, while the model demonstrates high classification accuracy on visually distinctive categories, fine-grained urban and natural classes with overlapping spectral-textural signatures remain challenging. This suggests that additional strategies, such as domain-specific prompt tuning or integrating contextual information about surrounding land-use, could further enhance performance on RESISC45.

\subsection{AID}
\begin{figure}[H]
\centering
  \includegraphics[trim={0 0 0 0},clip, width=1.0\linewidth]{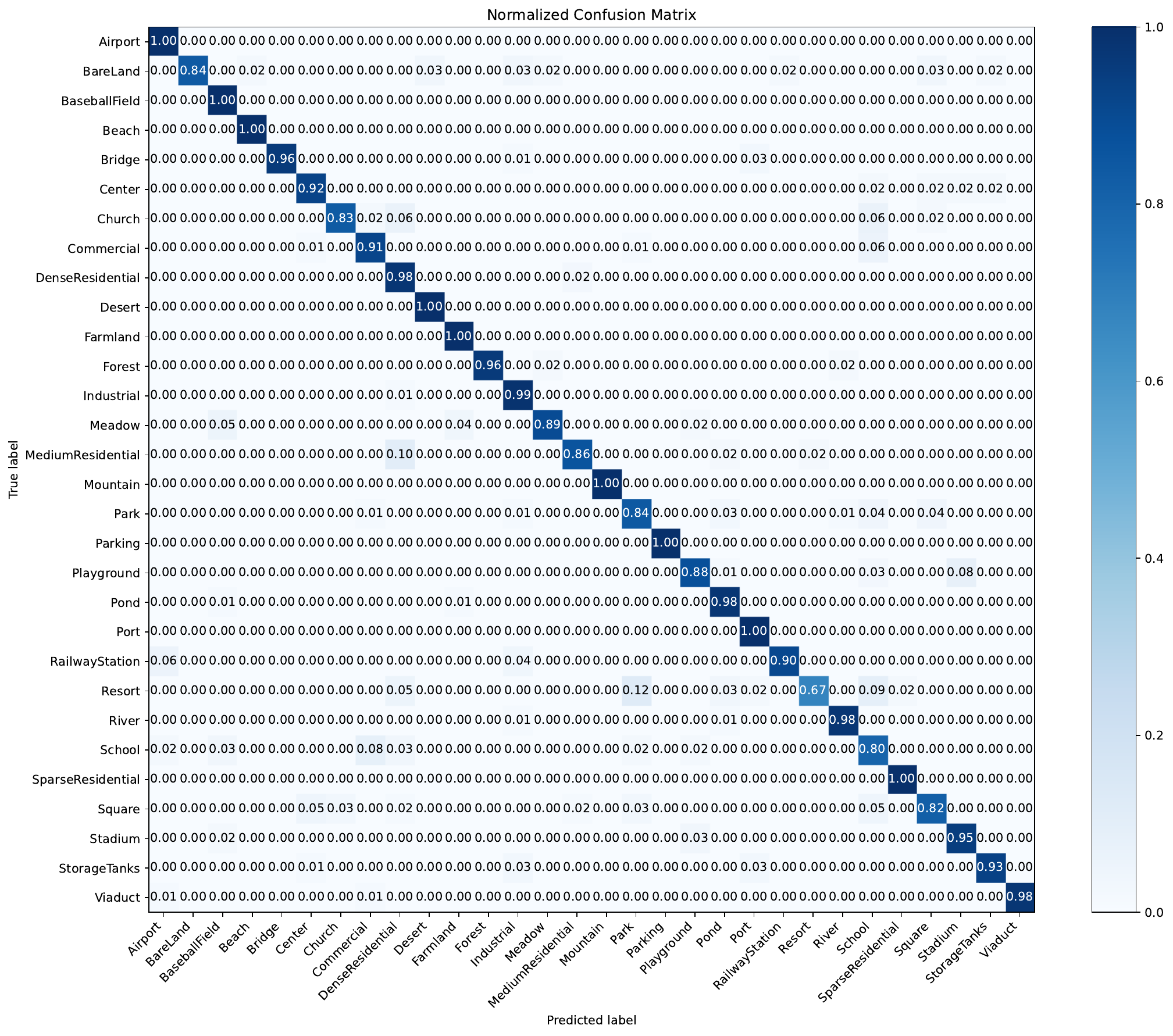}
  \caption{Normalized confusion matrix illustrating per-class performance of the best-performing configuration (CLIP-PromptSRC, 16-shot setting) on the AID dataset.}
  \label{fig:cmat_aid}
\end{figure}

The normalized confusion matrix for the AID dataset, presented in Figure~\ref{fig:cmat_aid} demonstrates strong classification performance across a broad range of scene categories, with most diagonal entries exceeding 0.85. Distinctive classes such as airport, baseball field, beach, bridge, commercial, dense residential, desert, farmland, forest, industrial, mountain, parking, pond, port, river, sparse residential, stadium, storage tanks, and viaduct achieve particularly high recognition rates ($\geq0.90$), highlighting the model’s ability to capture their characteristic spatial-structural and spectral signatures.

Despite this robustness, certain systematic confusions are evident. Bare land is occasionally misclassified as meadow, which can be explained by spectral similarities and vegetation sparsity in these categories. Church exhibits overlap with dense residential and school, likely due to architectural resemblance and contextual surroundings in urban layouts. Misclassifications also appear among residential categories: medium residential is sometimes confused with dense residential, reflecting gradual density transitions and overlapping textural cues. Additionally, railway station shows confusion with industrial and airport, which may result from similarities in large structural layouts and adjacent open areas.

Overall, the confusion matrix highlights the model’s strong discriminative capacity in identifying highly distinctive classes while revealing challenges in separating categories with subtle semantic or visual overlaps, particularly in urban and vegetation-related scenes. These results suggest that targeted prompt refinement or incorporating auxiliary spatial-contextual cues could further improve fine-grained discrimination in the AID dataset.

\subsection{RSSCN7}
\begin{figure}[H]
\centering
  \includegraphics[trim={0 0 0 0},clip, width=0.7\linewidth]{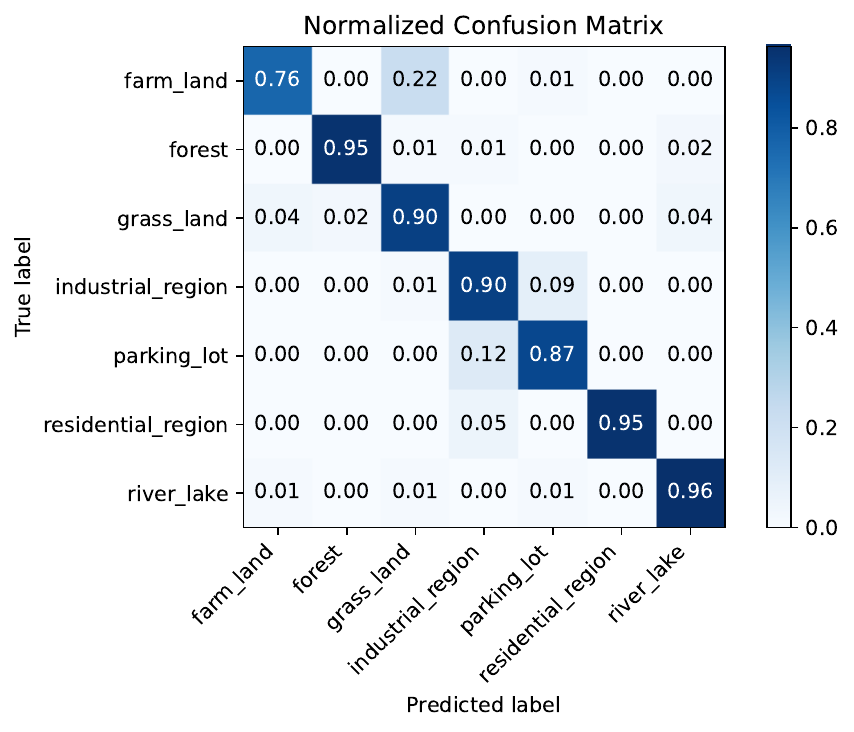}
  \caption{Normalized confusion matrix illustrating per-class performance of the best-performing configuration (CLIP-MaPLe, 16-shot setting) on the RSSCN7 dataset.}
  \label{fig:cmat_rsscn7}
\end{figure}

The normalized confusion matrix for the RSSCN7 dataset, presented in Figure~\ref{fig:cmat_rsscn7} shows that the model achieves consistently high per-class accuracy, with most diagonal entries above 0.90. Particularly strong recognition is observed for forest, residential region and river/lake, where the model exhibits near-perfect discrimination. Similarly, industrial region, grassland, and parking lot demonstrate high reliability, confirming the robustness of the learned representations for natural landscapes and well-structured human-made areas.

Some misclassification patterns emerge in categories with partial visual overlap. Farm land is occasionally predicted as grassland, likely due to shared vegetation textures and varying agricultural practices that create ambiguity in aerial imagery. Industrial region shows confusion with parking lot, which can be explained by the presence of large paved areas, buildings, and contextual similarities across these urban-related classes.

In general, the results on RSSCN7 highlight the strong generalizability of the model in both natural and urban categories, with classification errors concentrated in semantically or structurally adjacent classes. These findings suggest that while the model provides robust performance in this relatively small dataset, further improvements in separating agricultural and urban-industrial categories could be achieved through refined prompt design or the inclusion of auxiliary spatial-contextual cues.

\subsection{Optimal-31}
\begin{figure}[H]
\centering
  \includegraphics[trim={0 0 0 0},clip, width=1.0\linewidth]{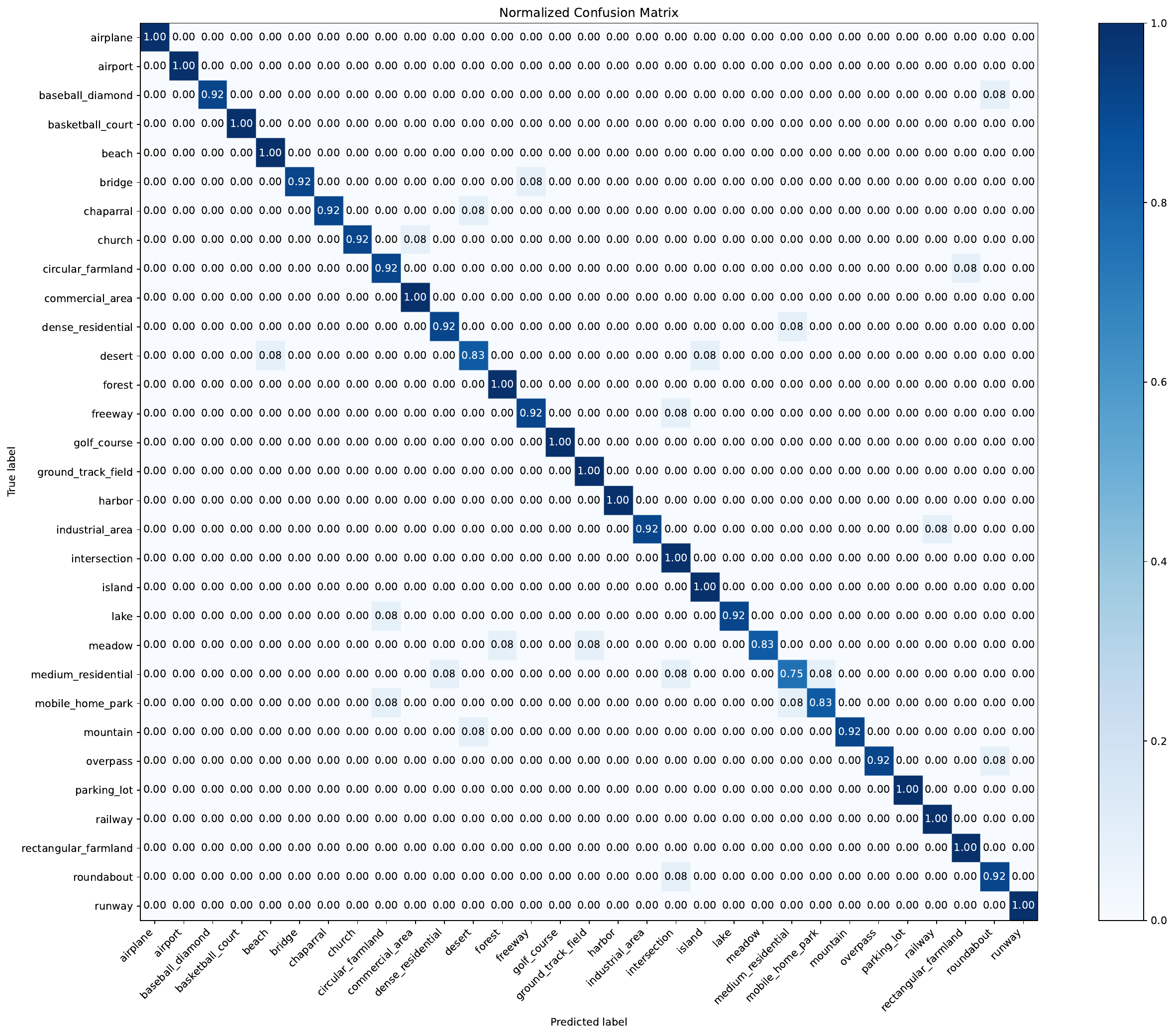}
  \caption{Normalized confusion matrix illustrating per-class performance of the best-performing configuration (CLIP-PromptSRC, 16-shot setting) on the Optimal-31 dataset.}
  \label{fig:cmat_optimal31}
\end{figure}

The normalized confusion matrix for the Optimal-31 dataset, presented in Figure~\ref{fig:cmat_optimal31} shows that the model achieves robust classification across a wide range of scene categories, with diagonal values generally above 0.85. Several categories, such as airplane, airport, basketball court, beach, commercial area, forest, golf course, ground track field, harbor, intersection, island, parking lot, railway, rectangular farmland, and runway are recognized with perfect accuracy, reflecting the strong discriminative power of the model for visually distinctive classes.

However, certain systematic confusions are observed in semantically or structurally related categories. Baseball field is occasionally misclassified as roundabout, both of which share visual similarities with sports and road-layout structures. Bridge exhibits confusion with freeway, reflecting overlapping linear structural patterns in overhead imagery. Church is sometimes predicted as commercial area, likely due to architectural resemblance or shared urban context. Residential classes such as dense residential show confusion with medium residential, highlighting the difficulty of separating high-density urban features with strong structural overlaps. In natural categories, desert occasionally overlaps with beach and island, suggesting challenges in disambiguating vegetation sparsity and transitional landscapes.

Overall, the Optimal-31 results highlight the model’s strong performance on well-separated scene types, while errors concentrate among categories with high visual or semantic proximity. These include small-scale man-made structures and transitional natural environments. Improvements could potentially be achieved through finer-grained prompt designs that emphasize contextual differences or through hierarchical classification strategies that first separate broader land-cover types before resolving subcategories.

\subsection{SIRI-WHU}
\begin{figure}[H]
\centering
  \includegraphics[trim={0 0 0 0},clip, width=0.7\linewidth]{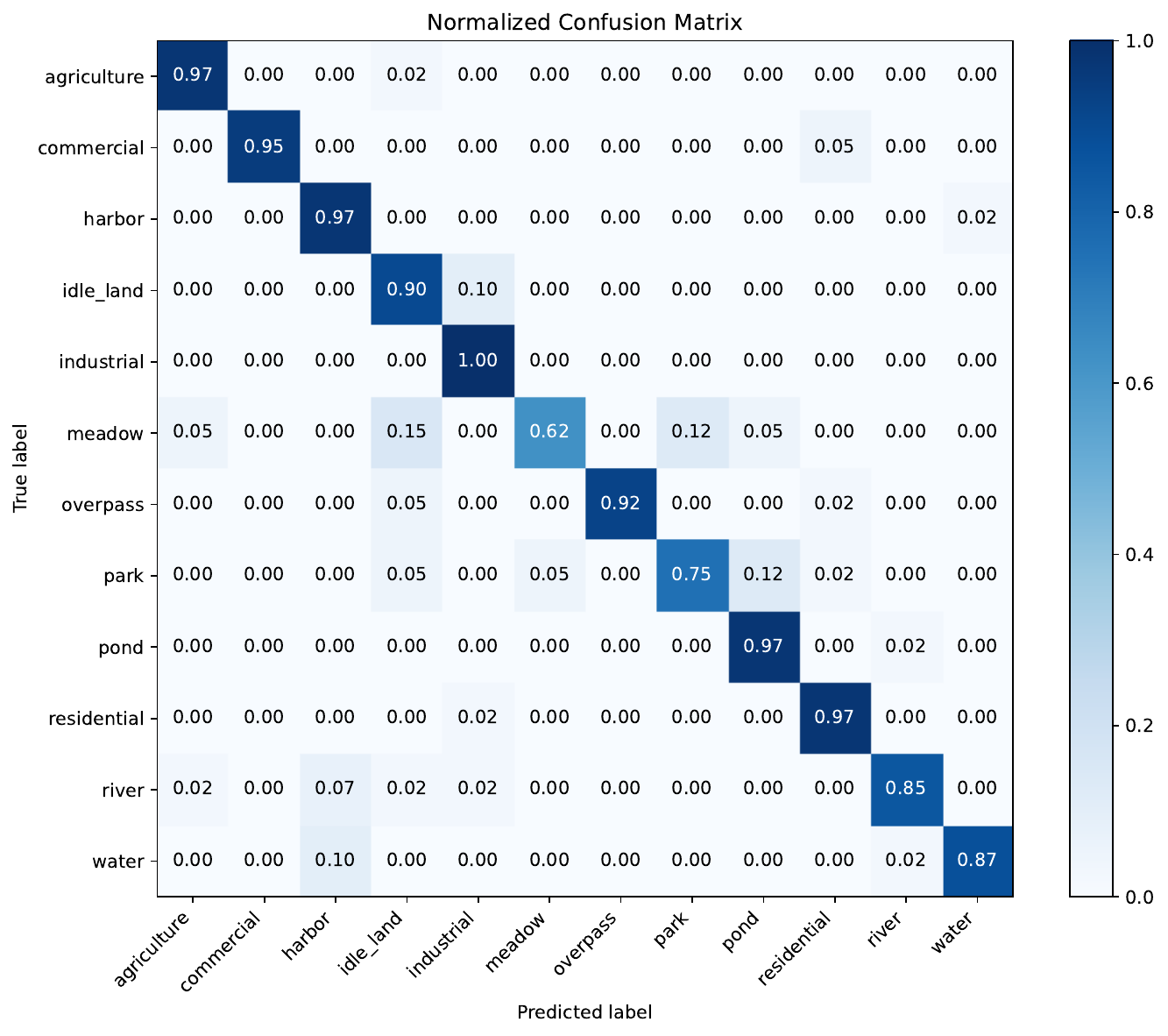}
  \caption{Normalized confusion matrix illustrating per-class performance of the best-performing configuration (CLIP-PromptSRC, 16-shot setting) on the SIRI-WHU dataset.}
  \label{fig:cmat_siriwhu}
\end{figure}

The normalized confusion matrix for the SIRI-WHU dataset, presented in Figure~\ref{fig:cmat_siriwhu} demonstrates that the model achieves strong overall performance, with most classes exhibiting per-class accuracies above 0.85. Distinctive categories such as agriculture, commercial, harbor, industrial, pond, and residential are recognized with near-perfect reliability, reflecting their unique spatial-structural and spectral characteristics in overhead imagery.

Despite this robustness, several systematic confusions appear. Meadow is sometimes misclassified as idle land or park, which is expected given shared vegetation textures and differences in land-use annotation. Similarly, park overlaps with meadow and pond, reflecting the challenge of distinguishing small-scale water bodies surrounded by vegetation. The residential category shows limited confusion with industrial, likely due to the similarity in urban textures and mixed-use layouts. River is occasionally misclassified as harbor, again highlighting the difficulty of separating water categories that differ primarily in size and context.

Overall, the results on SIRI-WHU confirm the model’s strong ability to discriminate large-scale and structurally distinct categories while revealing challenges in fine-grained natural vegetation and water-related classes. These confusions suggest that while CLIP-PromptSRC with 16 shots effectively adapts to this dataset, further gains could be achieved through refined prompts emphasizing land-use semantics (e.g., scale cues for water bodies) or incorporating spatial-contextual priors to better separate overlapping vegetation and urban categories.

\subsection{CLRS}
\begin{figure}[H]
\centering
  \includegraphics[trim={0 0 0 0},clip, width=1.0\linewidth]{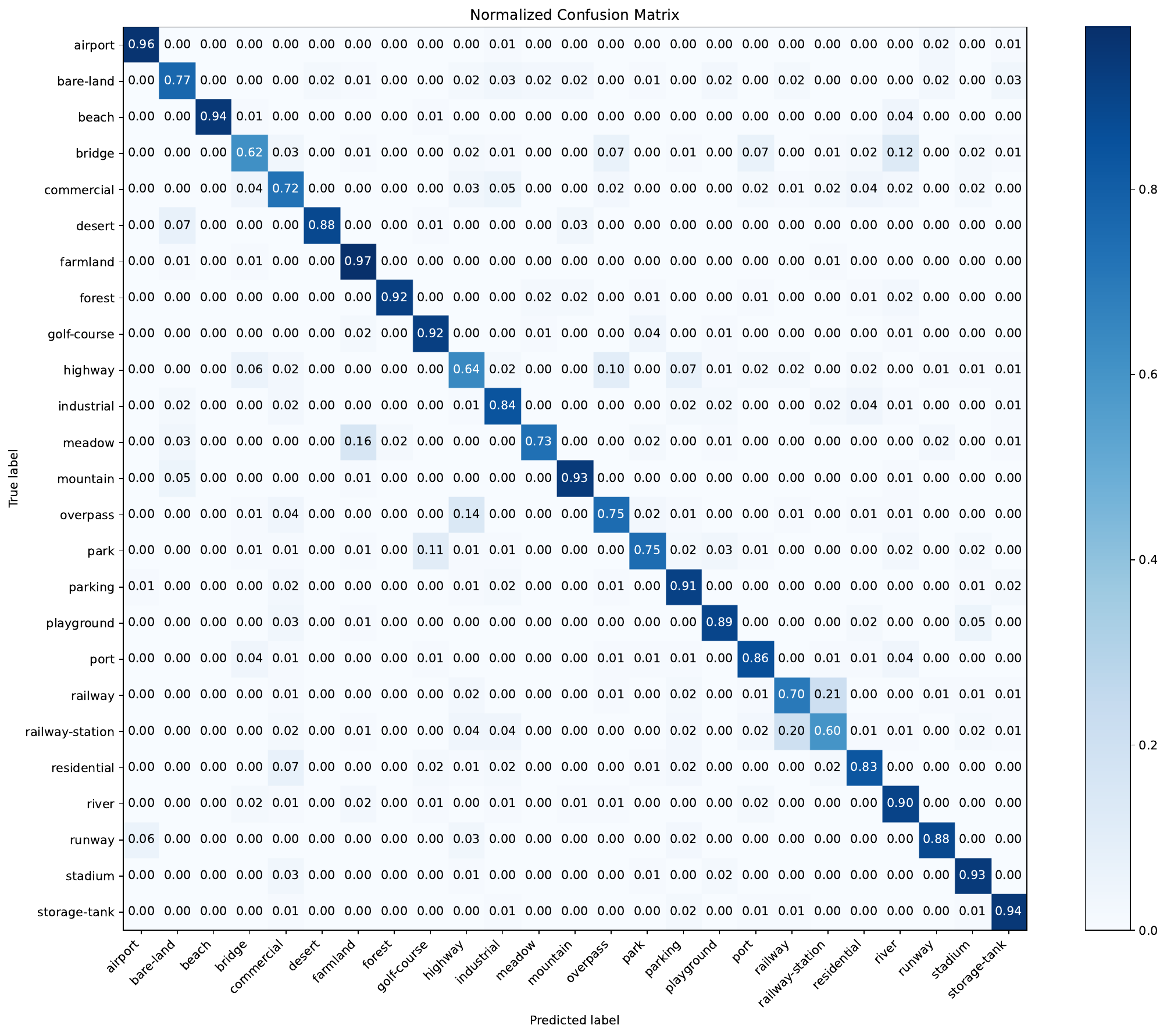}
  \caption{Normalized confusion matrix illustrating per-class performance of the best-performing configuration (CLIP-PromptSRC, 16-shot setting) on the CLRS dataset.}
  \label{fig:cmat_clrs}
\end{figure}

The normalized confusion matrix for the CLRS dataset, presented in Figure~\ref{fig:cmat_clrs} shows a more challenging scenario compared to other benchmarks, with several classes achieving strong performance but others exhibiting notable confusion. High per-class accuracy is observed for categories such as airport, beach, farmland, forest, golf-course, mountain, stadium, and storage tank, demonstrating the model’s ability to reliably capture distinctive structural and spectral features of these scenes.

However, substantial misclassifications are evident in several semantically related or visually overlapping categories. Bridge is frequently confused with overpass, river, and port, likely due to shared linear structures and contextual surroundings in urban areas. Commercial shows overlap with residential and industrial, reflecting the difficulty in disambiguating densely built environments. Furthermore, railway is often misclassified as railway station, indicating that the model struggles to separate these semantically proximate classes. Other confusions include playground with stadium, pointing to challenges in distinguishing certain small-scale man-made and natural environments.

Overall, the CLRS results highlight both the strengths and limitations of the best-performing model (CLIP-PromptSRC, 16 shots). While highly distinctive categories are classified with high reliability, categories with subtle semantic differences, particularly in residential and transportation-related scenes, remain problematic. These findings suggest that incorporating finer-grained prompts or hierarchical classification strategies (e.g., first distinguishing natural vs. man-made scenes, then resolving subcategories) could help mitigate these errors.

\subsection{MLRSNet}
\begin{figure}[H]
\centering
  \includegraphics[trim={0 0 0 0},clip, width=1.0\linewidth]{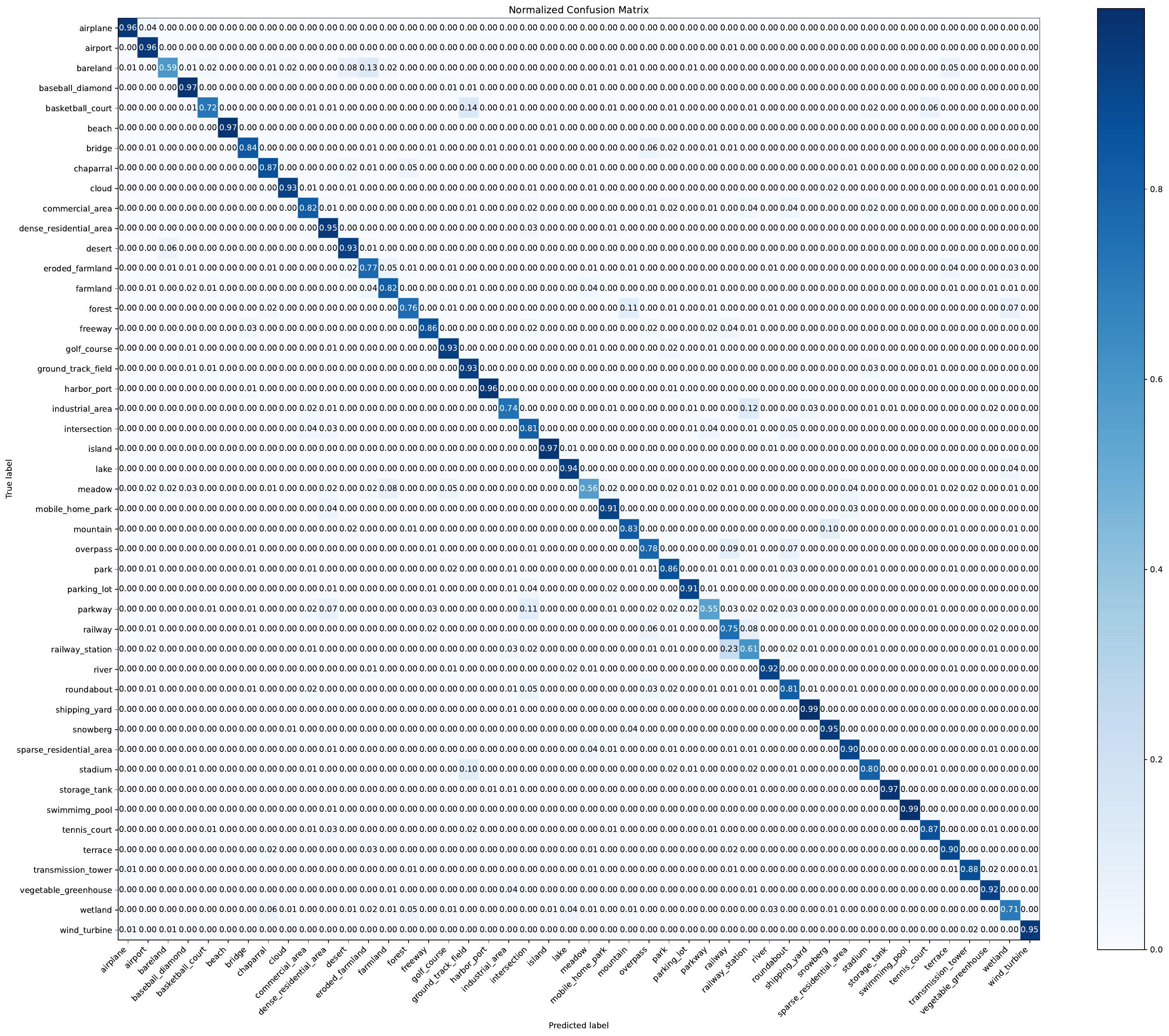}
  \caption{Normalized confusion matrix illustrating per-class performance of the best-performing configuration (CLIP-PromptSRC, 16-shot setting) on the MLRSNet dataset.}
  \label{fig:cmat_mlrsnet}
\end{figure}

The normalized confusion matrix for the MLRSNet dataset, presented in Figure~\ref{fig:cmat_mlrsnet} highlights both the strengths and limitations of the best-performing model (CLIP-PromptSRC, 16 shots). Several categories are recognized with near-perfect reliability, including airplane, airport, baseball diamond, beach, harbor\_port, island, shipping\_yard, swimming pool, and storage tank. These results confirm that the model effectively captures strong spectral-spatial signatures of both natural landscapes and large-scale man-made structures.

In contrast, some categories remain considerably more challenging. The weakest performance is observed for parkway, bareland, meadow, and railway\_station. Analysis of off-diagonal entries shows that these classes are often confused with semantically or structurally similar ones. Railway and railway station exhibit strong mutual confusion, reflecting their close semantic and spatial relationship. Meadow is frequently misclassified as farmland, due to overlapping vegetation textures. Parking lot and parkway are confused with other urban categories such as commercial area or industrial area, where large paved surfaces dominate. Bareland overlaps with desert and eroded farmland, which share sparse vegetation and similar spectral characteristics.

Overall, the MLRSNet results underscore the model’s robustness in classifying visually distinctive and structurally consistent categories, while fine-grained separation of semantically related land-cover types remains problematic. Errors concentrate in urban, transitional vegetation, and transportation-related classes, where subtle differences in density, layout, or context can blur category boundaries. Addressing these limitations may require incorporating hierarchical classification schemes (e.g., first separating natural vs. urban vs. transportation) or context-aware prompt designs that emphasize discriminative cues such as scale, density, or functional semantics.

\end{appendices}

\bibliographystyle{unsrt}
\bibliography{template}% common bib file

\end{document}